\documentclass[sigconf]{acmart}

 \setcopyright{none} 

\acmConference[CCS '20] {2020 ACM SIGSAC Conference on Computer and Communications Security}{November 9--13, 2020}{Virtual Event, USA}
\acmDOI{10.1145/3372297.3417880}

\usepackage[utf8]{inputenc} 
\usepackage[T1]{fontenc}    
\usepackage{hyperref}       
\usepackage{url}            
\usepackage{booktabs}       
\usepackage{amsfonts}       
\usepackage{amsthm}
\usepackage{nicefrac}       
\usepackage{microtype}      
\usepackage{xcolor,colortbl}
\usepackage{paralist}
\usepackage{algorithm}
\usepackage[noend]{algpseudocode}
\usepackage{multirow}
\usepackage{listings}
\usepackage{graphicx}
\usepackage{censor}
\usepackage{siunitx}
\usepackage{subcaption}
\usepackage{balance}
\usepackage{adjustbox}
\usepackage{dblfloatfix}

\lstdefinelanguage{JavaScript}{
  keywords={typeof, static, new, begin, end, struct, char, void, unsigned, long, const, privilege_enclave, int, true, false, catch, function, return, null, catch, switch, var, if, in, while, do, else, case, break},
  keywordstyle=\color{blue},
  ndkeywords={class, export, boolean, throw, implements, import, this},
  ndkeywordstyle=\color{darkgray}\bfseries,
  identifierstyle=\color{black},
  sensitive=false,
  comment=[l]{//},
  morecomment=[s]{/*}{*/},
  commentstyle=\color{purple}\ttfamily\bfseries,
  stringstyle=\color{red}\ttfamily,
  morestring=[b]',
  morestring=[b]"
}

\lstdefinestyle{JavaScript}{
    language={JavaScript},
    moredelim=**[is][\btHL]{`}{`},
    moredelim=**[is][{\btHL[fill=green!30,]}]{@}{@},
}

\lstdefinestyle{nonumbers}
{numbers=none}

\definecolor{Blue3}{HTML}{0000CD}
\definecolor{Green4}{HTML}{008B00}
\definecolor{Red3}{HTML}{CD0000}
\definecolor{orange}{rgb}{0.8, 0.47, 0.196}
\definecolor{Red}{rgb}{1, 0, 0}

\newcommand\sizeof[1]{\left|#1\right|}

\newcommand\drank{\mathit{DR}}
\newcommand\dscore[2]{\mathit{DS}_{#1}^{#2}}

\newcommand\perpl{\mathit{perp}}

\newcommand\drelscore[2]{\widetilde{\mathit{DS}}_{#1}^{#2}}

\newcommand{\tokens}{T}

\usepackage{amsmath,amsfonts,bm}

\def\eqref#1{equation~\ref{#1}}

\def\1{\bm{1}}

\def\eps{{\epsilon}}

\DeclareMathAlphabet{\mathsfit}{\encodingdefault}{\sfdefault}{m}{sl}
\SetMathAlphabet{\mathsfit}{bold}{\encodingdefault}{\sfdefault}{bx}{n}

\begin{document}

\fancyhead{}

\title[Analyzing Information Leakage of Updates to Natural Language Models]
      {Analyzing Information Leakage of Updates\texorpdfstring{\\}{} to Natural Language Models}\titlenote{ \copyright 2020 Copyright held by the author(s). This is the author's version of the work. It is posted here for your personal use. Not for redistribution.
 The definitive version was published in CCS '20: Proceedings of the 2020 ACM SIGSAC Conference on Computer and Communications Security
October 2020, pp 363--375, \url{https://doi.org/10.1145/3372297.3417880}.}

\author{Santiago Zanella-B\'eguelin}
\email{santiago@microsoft.com}
\affiliation{Microsoft}

\author{Lukas Wutschitz}
\email{luwutsch@microsoft.com}
\affiliation{Microsoft}

\author{Shruti Tople}
\email{shruti.tople@microsoft.com}
\affiliation{Microsoft}

\author{Victor R\"uhle}
\email{virueh@microsoft.com}
\affiliation{Microsoft}

\author{Andrew Paverd}
\email{andrew.paverd@microsoft.com}
\affiliation{Microsoft}

\author{Olga Ohrimenko}
\email{oohrimenko@unimelb.edu.au}
\affiliation{University of Melbourne}
\authornote{Work done in part while at Microsoft.}

\author{Boris K\"opf}
\email{boris.koepf@microsoft.com}
\affiliation{Microsoft}

\author{Marc Brockschmidt}
\email{mabrocks@microsoft.com}
\affiliation{Microsoft}

\renewcommand{\shortauthors}{Zanella-B\'eguelin, Wutschitz, Tople, R\"uhle, Paverd, Ohrimenko, K\"opf, and Brockschmidt}

\begin{abstract}
  To continuously improve quality and reflect changes in data, machine
  learning applications have to regularly retrain and update their
  core models.
  We show that a differential analysis of language model snapshots before and
  after an update can reveal a surprising amount of detailed information about
  changes in the training data.
  We propose two new metrics---\emph{differential score} and
  \emph{differential rank}---for analyzing the leakage due to updates
  of natural language models. We perform leakage analysis using these
  metrics across models trained on several different datasets using
  different methods and configurations.
  We discuss the privacy implications of our findings, propose
  mitigation strategies and evaluate their effect.
\end{abstract}

\maketitle

\section{Introduction}
\label{sec:introduction}
Over the last few years, deep learning has made sufficient progress to
be integrated into intelligent, user-facing systems, which means that
machine learning models are now part of the software development
lifecycle. As part of this cycle, models are regularly updated to
accommodate three different scenarios:
\begin{itemize}
\item \emph{data update}, to improve performance when new and more
  data becomes available;
\item \emph{data specialization}, to fine-tune a model on a
  specific dataset, or to handle distributional shift as usage
  patterns change; or
\item \emph{data deletion}, to respect requests for removal of
  users' data.
\end{itemize}

Motivated by these scenarios, we study privacy implications for text
data that is added (or removed) during retraining of generative
natural \underline{l}anguage \underline{m}odels (LMs). Specifically,
we consider an adversary with access to multiple snapshots of a
model and wishes to learn information about differences in the data
used to train them.
This threat model is motivated by the combination of three factors:
\begin{inparaenum}
\item the current trend to fine-tune pretrained public high-capacity
  LMs to smaller private datasets;
\item the established ability of such LMs to memorize
  out-of-distribution training samples~\citep{secret-sharer}; and
\item the widespread deployment of LMs to end-user systems (e.g.,
  predictive keyboards on smartphones), allowing adversaries to
  analyze them in detail.
\end{inparaenum}
For the informed reader, we discuss the relationship between this
threat model and other attacks against privacy and defenses like
differential privacy later on in Section~\ref{sec:adversary_model}.


We show that data that is added or removed between model updates can
be extracted in this threat model, having severe implications for
deploying machine learning models trained on private data.
Some of the implications are counter-intuitive: for example, honoring
a request to remove a user's data (as per GDPR) from the training
corpus can mean that their data becomes exposed by releasing an
updated model trained without it.
Similarly, fine-tuning a public snapshot of a high-capacity model
(e.g., BERT~\citep{devlin2018bert} or
GPT-2~\citep{radford2019language}) with data from a single
organization exposes this additional data to anyone with access
to both the fine-tuned model and the original public model (e.g.,
employees of this organization).
%


In order to extract information about the difference in the data used
to train two language models, we develop a novel notion of
\emph{differential score}.
The differential score of a token sequence captures the difference
between the probabilities assigned to it by the two models.
The intuition is that sequences with higher differential
scores are likely to have been added during model updates.
We devise an algorithm based on beam search to efficiently
identify such token sequences, even if the individual models assign
low probability to them. This allows us to recover
information about the difference between the datasets used for
training \emph{without} any background knowledge of their contents or
distribution.

When given \emph{some} background knowledge, the advantage of having
access to two model snapshots becomes crisper.
For example, we train a recurrent neural network (RNN) on 20M tokens
of general Reddit comments, and update it by retraining it on these
comments plus 25K tokens from 940 messages of the
\texttt{talk.politics.mideast} newsgroup.
When prompted with the word ``Turkey'', our algorithm produces
``Turkey searched an American plane'' as the 2$^\textrm{nd}$ most
likely result, although this phrase occurs only 6 times in newsgroup
messages and none in Reddit comments (i.e., $<0.000002\%$ of the training data).
An equivalent search using only the updated network does not produce
this sentence among the top 10,000 results; it would take the longer
prompt ``Turkey searched an'' for this phrase to surface to the top
100 results.



We use differential score to experimentally study the
effect of updates in the three scenarios mentioned above. As a proxy
for the update dataset, we use synthetically generated sentences (or
\emph{canaries}) and real-world sentences from newsgroup
messages. Using both canaries and real-world data, we analyze the
effect on attacks recovering information from the update dataset of
\begin{inparaenum}
\item different training types for updates, ranging from
  retraining a model from scratch with an updated dataset to
  fine-tuning as is common for modern high-capacity language models;
\item the proportion of private and public data used for the update;
  and
\item an adversary's background knowledge.
\end{inparaenum}
For robustness, we consider datasets of different sizes on both RNNs
as well as modern transformer architectures.

\paragraph{Summary of Contributions}
We present the first systematic study of the privacy implications of
releasing snapshots of language models trained on overlapping data.
%
%
Our results validate that model updates pose a substantial
risk to content added to or removed from training data in terms of information
leakage.
%
Our key findings are:
\begin{asparaitem}
\item By comparing two models, an adversary can extract specific
  sentences or fragments of discourse from the difference between the
  data used to train them.  This does not require any information
  about the training data or the model architecture and is possible
  even when the change to the data is as small as 0.0001\% of the
  original dataset. Smaller changes become exposed when given partial
  knowledge about the data.
\item We show that analyzing two model snapshots reveals substantially
  more about the data that was added or removed than considering only
  a single snapshot at a time, as in~\cite{secret-sharer}.
\item Adding or removing additional non-sensitive training data
  between model updates is not a reliable mitigation.
\item
  Training with differential privacy mitigates the attack, but incurs
  substantial computational cost and reduces the utility of the
  trained models.
\item Restricting access to the model and only outputting
  a subset of prediction results is a promising mitigation as it
  reduces the effectiveness of our attack without reducing utility of
  the model.
\end{asparaitem}
These findings apply to models fine-tuned on a smaller
dataset, as well as models retrained on the union of original
and new data.

\paragraph{Structure of the Paper}
We provide background on language models and describe our adversary
model and attack scenarios in the next section.
We define the notion of differential score and describe how to
efficiently approximate it in Section~\ref{sec:methodology}.
In Section~\ref{sec:experiments} we describe our experiments to
analyze the effect of different factors on leakage. In
Section~\ref{sec:leaksource} we investigate the source of leakage in
model updates, e.g., by comparing with leakage from access to only a
single model.
Finally, we consider mitigation strategies in
Section~\ref{sec:mitigations}, before describing related work and
concluding.

\section{Preliminaries}
\label{sec:adv_and_scenarios}
\subsection{Generative Language Models}

We consider machine learning models capable of generating natural
language. These models are used in a variety of applications,
including automatic caption generation, language translation, and
next-word prediction.  Generative language models usually operate on a
fixed set of known tokens $\tokens$ (often referred to as the model's
\emph{vocabulary}) and are \emph{autoregressive}, modeling the
probability $p(t_1 \ldots t_n)$ of a sequence of tokens $t_1 \ldots
t_n \in \tokens^n$ as the product of the per-token probabilities
conditional on their prefix $p(t_i \mid t_1 \ldots t_{i-1})$, i.e.,
\begin{displaymath}
  p(t_1 \ldots t_n) = \prod_{1 \leq i \leq n} p(t_i \mid t_1 \ldots t_{i-1}) \,.
\end{displaymath}
Training an autoregressive generative language model $M$
requires learning a function (which we also refer to as $M$) that maps
token sequences of arbitrary length to a probability distribution over
the vocabulary $\tokens$, modeling the likelihood of each token to
appear next.
We use $M(t_{<i})$ to denote the probability distribution over
tokens computed by model $M$ after reading the sequence $t_1 \dots
t_{i-1} \in \tokens^*$, and $M(t_{<i})(t_i)$ to denote the probability
of a specific token $t_i$.

Given such a model $M$, a simple predictive screen keyboard can be
implemented by feeding $M$ the words typed so far (e.g., from the
start of the current sentence) and displaying the, say, three most
likely tokens as one-tap options to the user.

A variety of different architectures exist for the generation of
natural language using machine learning models.
The most prominent are Recurrent Neural Networks (RNNs) using Long
Short-Term Memory~\cite{Hochreiter97} cells (or variants thereof) and
the more recent
Transformers~\cite{vaswani2017attention,radford2019language}.
These architectures differ substantially in how they implement the
modeling of the per-token probability distribution, but as our
experiments show, they behave nearly identically for the purposes
of our analysis.

Given a model architecture, a dataset $D \subseteq \tokens^*$ is
required as training data to obtain a concrete model.
We write $M_D$ to emphasize that a model was trained on a dataset $D$.
Throughout the paper, we use the standard measure of
\emph{perplexity} $\perpl_M(t_1 \ldots t_n) = p_M(t_1 \ldots
t_n)^{\frac{-1}{n}}$ of a model $M$ on test data $t_1 \ldots t_n$,
using the probability $p_M(t_1 \ldots t_n)$ assigned to the sequence
by model $M$.
Unlike the more familiar accuracy, which only captures the correctness
of the most probable choice, this metric captures models being
``almost right.'' Intuitively, perplexity can be thought as how
``surprised'' a model is by a next-word choice, and hence, lower
perplexity values indicate a better match between data and model.

\subsection{Adversary Model and Goals}
\label{sec:adversary_model}

Language models are regularly \emph{updated} for a variety of reasons,
either by adding and/or removing data from the training set. We use
the term \emph{model update} to refer to any update in the parameters
of the model caused by training on different data.  This is distinct
from an update to the model architecture, which changes the number or
use of parameters.  Each update creates a new version of the model,
which we refer to as a \emph{snapshot}.

We consider an adversary that has concurrent query access to two
snapshots, $M_D$ and $M_{D'}$, of a language model trained on datasets
$D$ and $D'$ respectively, where $D \subsetneq D'$. We write $M$, $M'$
as shorthand for $M_D$, $M_{D'}$.
The adversary can query the snapshots with any sequence $s\in
\tokens^*$ and observe the corresponding probability distributions
$M(s)$ and $M'(s)$.
The adversary's goal is to infer information about training data
points in $D' \setminus D$, the difference between $D$ and $D'$. In
the best case, an adversary would recover exact training points.
We refer to an adversary who has access to two snapshots of the model
as \emph{a snapshot attacker}.

\paragraph{Relationship to other attacks on training data.}

Snapshot attacks are {\em reconstruction attacks}~\citep{DBLP:journals/corr/abs-1904-01067} against the updated model, as the goal is to recover data points in the dataset used for the update, given the
original model as auxiliary information.

The goal of \emph{membership inference attacks}~\citep{DBLP:conf/sp/ShokriSSS17,Salem:NDSS19} is weaker in that they
only aim to determine whether a given point was present in the dataset used to
train a model.
However, the differential score of a phrase (which we use for reconstruction) can also serve as a signal for
inferring membership in the update dataset. We leave an evaluation of this approach to future work. 

Finally, \emph{model inversion attacks}~\citep{Fredrikson:2014,Fredrikson:2015}
repurpose a model to work \emph{backwards}, inferring unknown
attributes of individuals given known attributes and a target
prediction. Individuals need not be present in the training data, and
results are aggregate statistics rather than information about
specific training points. See Section~\ref{sec:related} for a more in-depth discussion of related attacks.

\paragraph{Relationship to differential privacy.}

Differential privacy~\citep{privacybook} guarantees that a model does
not leak significant information about any specific training point.
A differentially private model also guarantees \emph{group} privacy,
with a bound on the contribution of a group of training points that
degrades linearly with the group size.
%
%
A differentially private model that provides meaningful protection for
a group of $|D' \setminus D|$ training points would hence protect against
snapshot attacks on $M_D,M_{D'}$.
However, this also implies that $M_{D'}$ cannot be significantly more
useful (e.g. more accurate) than $M_D$.
Our experiments in Section~\ref{sec:mitigations} confirm this
intuition, and show that a large privacy budget is needed for the
updated model to gain in utility, so that in practice differential
privacy provides an empirical mitigation rather than a strong formal
guarantee.


\subsection{Analysis Scenarios}
\label{sec:scenarios}

To guide our analysis, we focus on three concrete scenarios in which
an adversary can gain concurrent access to two (or more) snapshots of
a language model.

\paragraph{Data Updates}
Many applications require language models that reflect recent patterns
in language use.
For example, a predictive keyboard on a mobile device requires regular
updates to suggest terms that have become more common recently (e.g.,
following news trends or internet memes).
To achieve this, vendors often regularly retrain an (otherwise
unchanged) model on an updated dataset, for example by simply adding
more recent data to the training dataset.
In such cases, an adversary can easily gain access to two snapshots
$M_D$ and $M_{D'}$ with $D \subsetneq D'$ and may be interested in
learning details about the update $D' \setminus D$.
We show that we can extract entire sentences from this difference
by comparing $M_D$ and $M_{D'}$, revealing not only aggregate user
behavior, but specific conversations.

\paragraph{Data Specialization}
Some applications with little task-specific data build on top of
generic, pretrained high-capacity language models such as
GPT-2~\cite{radford2019language}.
In such settings, training starts from the pretrained model, but then
uses a significantly smaller private dataset.
As an example, an organization could simply use a publicly available
off-the-shelf language model to create an email authoring
autocompletion system.
However, by additionally training the model with some historical email
data, it can be adapted to organization-specific terms, acronyms and
concepts. In such a scenario, if an adversary can gain access to the
specialized model $M'$, they can easily also obtain the (publicly
available) model $M$ used as a basis.
We show that by treating these as different snapshots of the same
model, the adversary can extract parts of the private dataset used for
specialization.

\paragraph{User Data Deletion}
Art.~17 of GDPR~\cite{GDPR} Right to erasure (``right to be forgotten'') gives
data owners the right to request erasure of their personal data from a
party who has collected and processed
it.
%
Language models trained on emails, text messages, or other
user-generated content may contain personal information that a user
can request to delete.
The data collector would be required to delete the user's data and retrain any models in which it had been used.
%
In many cases, these models may have already been released either to the public or to
other users via services provided by the data collector (e.g., text
prediction and auto-correct services in text editors and mobile
keyboards).

This scenario falls into our adversary setting, albeit in reverse
chronological order. Here the dataset $D'$ contains the data that
will be deleted, whilst $D$ does not (i.e., the difference $D'
\setminus D$ represents the user's data).  With access to $M_D$ and
$M_{D'}$, the attacker can attempt to infer the user's data.  Even if
the retrained model overwrites the old model, it may not be possible
to erase all instances of the old model simultaneously.  For example,
some users may be slow to download the new version or the old model
may have been copied by other parties.

Naturally, this scenario can be extended to other settings where data
is deleted between model updates. This scenario raises an interesting
question on whether deletion of data is in the user's best interest or
if it makes their data more susceptible to  leakage.

\section{New Metrics}
\label{sec:methodology}
We introduce two metrics called differential rank and differential
score to analyze data exposure between two snapshots of a generative
language model.

\subsection{Differential Score and Differential Rank}

We aim to identify token sequences whose probability differs most
between models~$M$ and $M'$. Intuitively, such sequences are most
likely to be related to the differences between their corresponding
training datasets~$D$ and~$D'$.

To capture this notion formally, we define the {\em differential
  score} $(\dscore{}{})$ of token sequences, which is simply the sum
of the differences of (contextualized) per-token probabilities.
We also define a \emph{relative} variant $\drelscore{}{}$ based on the
relative change in probabilities, which we found to be more robust
w.r.t.~the \emph{noise} introduced by different random initializations
of the models $M$ and $M'$.

\begin{definition}
\label{def:dscore}
Given two language models $M, M'$ and a token sequence $t_1 \ldots t_n
\in \tokens^*$, we define the {\em differential score} of a token as
the increase in its probability and the {\em relative differential
  score} as the relative increase in its probability.  We lift these
concepts to token sequences by defining
\begin{align*}
  \dscore{M}{M'}(t_1\dots t_n)    &= \sum_{i=1}^n M'(t_{<i})(t_i) - M(t_{<i})(t_i) \, ,\\
  \drelscore{M}{M'}(t_1\dots t_n) &= \sum_{i=1}^n \frac{M'(t_{<i})(t_i) - M(t_{<i})(t_i)}{M(t_{<i})(t_i)} \, .
\end{align*}
\end{definition}

The differential score of a token sequence is best interpreted
relative to that of other token sequences. This motivates ranking
sequences according to their differential score.
\begin{definition}
\label{def:drank}
We define the {\em differential rank} $\drank(s)$ of $s \in
\tokens^\ast$ as the number of token sequences of length $\sizeof{s}$
with differential score higher than $s$.
\begin{displaymath}
  \drank(s) =
  \sizeof{\left\{
      s' \in \tokens^{\sizeof{s}} \left| \dscore{M}{M'}(s') > \dscore{M}{M'}(s) \right.
    \right\}} \,.
\end{displaymath}
\end{definition}

The lower the differential rank of a sequence, the more the sequence
is exposed by a model update, with the most exposed sequence having
rank 0.

\subsection{Approximating Differential Rank}
\label{sec:algo}

Computing the differential rank $\drank(s)$ of a sequence $s$ of
length $|s| = n$ requires searching a space of size
$\sizeof{\tokens}^{n}$.
To avoid exponential blow-up, we rely on
Algorithm~\ref{alg:beamsearch}, which approximates the differential
rank based on {\em beam search}.

At iteration $i$, the algorithm maintains a set $S$ of $k$ (called the
{\em beam width}) candidate sequences of length $i$ together with
their differential scores. The algorithm iterates over all
$k\cdot \sizeof{\tokens}$ single-token extensions of these sequences,
computes their differential scores, and keeps the $k$ highest-scoring
sequences of length $i+1$ for the next step. Eventually,
the search completes and returns the set $S$.

\begin{algorithm}[ht]
  \caption{Beam search for Differential Rank}\label{alg:beamsearch}
  \label{alg:beam_search}
  \begin{flushleft}
  \hspace*{\algorithmicindent} \textbf{In:} $M,M'$=models, $T$=tokens, $k$=beam width, $n$=length \\
  \hspace*{\algorithmicindent} \textbf{Out:} $S$=set of ($n$-gram, $\dscore{}{}$) pairs\\
  \end{flushleft}
  \begin{algorithmic}[1]
  \State {$S \leftarrow \{(\epsilon,0)\}$}  \Comment{Initialize with empty sequence $\epsilon$}
  \For{$i=1\dots n$}
  \State $S'\leftarrow\{(s\circ t, r+\dscore{M}{M'}(s)(t)) \mid (s,r)\in S, t\in T\}$
  \State $S\leftarrow \mathit{take(k,S')}$   \Comment{Take top $k$ items from $S'$}
  \EndFor
  \State \Return $S=\{(s_1,r_1),\dots,(s_k,r_k)\}$ such that $r_1\ge\dots \ge r_k$ 
  \end{algorithmic}
\end{algorithm}

Algorithm~\ref{alg:beamsearch} returns a set of token sequences $s$
and their differential score $r$. With this we can approximate the
differential rank $\drank(s)$ by the number of token sequences {\em
  in} $S$ with differential score higher than $s$. For large enough
beam widths this yields the true rank of $s$. For smaller widths,
the result is a {\em lower bound} on $\drank(s)$, as a search may miss
sequences with higher differential score.

\begin{proposition}
If Algorithm~\ref{alg:beamsearch} returns a set
\begin{displaymath}
S = \left\{ (s_1,r_1), \dots, (s_k,r_k) \right\} \textrm{ with } r_1 \ge \dots \ge r_k \, ,
\end{displaymath}
then $\dscore{M}{M'}(s_i)=r_i$ and $\drank(s_i)\ge i-1$.
\end{proposition}

\paragraph{Optimizing for Speed}
The beam width $k$ governs the trade-off between computational cost
and the precision of the approximation. In experiments, we found that
shrinking the beam width as the search progresses speeds up the search
considerably without compromising on the quality of results.
Typically, we use a beam width $\sizeof{T}$, which we halve at each
iteration. That is, we consider $\sizeof{T}/2$ candidate phrases of
length two, $\sizeof{T}/4$ sequences of length three, and so on.

\paragraph{Optimizing for Diversity}
Since the sequences returned by vanilla beam search typically share a
common prefix, we rely on \emph{group beam search} as a technique for
increasing diversity: we split the initial $\sizeof{T}$ one-token
sequences into multiple groups according to their differential score,
and run parallel beam searches extending each of the groups
independently. See~\cite{VijayakumarCSSL18} for more sophisticated
techniques for increasing diversity.

\section{Leakage Analysis}
\label{sec:experiments}
We use our new metrics to perform leakage analyses for various
datasets across various model update scenarios.
We first describe our benchmark datasets with their model
configurations and the model training scenarios we consider.
Then, we discuss research questions relevant to the analysis scenarios
described in Section~\ref{sec:scenarios}.
We then show experiments investigating these questions in detail,
first using synthetically generated canaries as a proxy for updates
where we can precisely control the differences between the datasets
used to create model snapshots, and then in a realistic setting, in
which we use a set of standard real-world datasets.

\subsection{Datasets and Models}
\label{sec:expsetup}

We consider three datasets of different size and complexity, matched
with standard model architectures whose capacity we adapted to the
data size and implemented in TensorFlow.%
\footnote{Source code and tools available at: \url{https://github.com/microsoft/language-privacy}}

Concretely, we use the Penn Treebank~\citep{marcus1993building} (PTB)
dataset as a representative of low-data scenarios, as the standard
training dataset has only around \num{900000} tokens and a vocabulary size
of \num{10000}. As the corresponding model, we use a two-layer recurrent
neural network using LSTM cells with 200-dimensional embeddings and
hidden states and no additional regularization (this corresponds to
the \emph{small} configuration of
\citet{DBLP:journals/corr/ZarembaSV14}).

Second, we use a dataset of Reddit comments with 20 million tokens
overall, of which we split off 5\% as validation set.
We use a vocabulary size of \num{10000}. We rely on two different model
configurations for this dataset, which allows us to understand the
impact of model size on information leakage using $\drank$ as a
metric.
\begin{enumerate}
\item a one-layer RNN using an LSTM cell with
512-dimensional hidden states and 160-dimensional embeddings. We employ
dropout on inputs and outputs with a keep rate of $0.9$ as
regularizer.
These parameters were chosen in line with a neural language model
suitable for next-word recommendations on resource-constrained mobile
devices.
\item a model based on the Transformer
architecture~\citep{vaswani2017attention} (more concretely, using the
BERT~\citep{devlin2018bert} codebase) with four layers of six
attention heads, each with a hidden dimension of~192.
\end{enumerate}

Finally, we use the Wikitext-103 dataset~\citep{merity2016pointer}
with 103 million training tokens as a representative of a big data
regime, using a vocabulary size of \num{20000}.
As the model, we employ a two-layer RNN with 512-dimensional LSTM
cells and token embedding size 512 and dropout on inputs and outputs
with a keep rate of $0.9$ as regularizer.
We combined this large dataset with this (relatively low-capacity)
model to test if our results still hold on datasets that clearly
require more model capacity than is available.

All models and their training are following standard best practices
for generative language models and represent common (simple) baselines
used in experiments on the used datasets.
This can be seen in the perplexity of the trained models on the
held-out test data, shown in \autoref{tbl:canary_results}, which is in
line with common test results.

\subsection{Implementing Model Updates}

Updated models can be created using different techniques, with
different applicability to the usage and analysis scenarios discussed
in Section~\ref{sec:scenarios}.

\paragraph{Retraining}
Given an updated dataset $D'$, a fresh model snapshot $M'$ can be
obtained by simply training a fresh model from scratch, which we refer
to as \emph{retraining}.
This also involves a fresh (random) initialization of the model
parameters, and in practice, retraining repeatedly on the
\emph{same} dataset will yield slightly different models.
\emph{Data deletion} requires updating a model to
eliminate the influence of some training data points at the request of
the data owner. This can be done by retraining a model after
pruning the data or, equivalently, using techniques with lower
computational cost~\citep{Ginart:NIPS2019,bourtoule2019machine}.

\paragraph{Continued Training}
In this approach, a fresh model snapshot $M'$ is obtained by taking an
existing model $M$ and continuing training it on additional data.
This is the core of the \emph{data specialization} scenario and
sometimes also used in \emph{data update} scenarios to avoid the
computational cost of training on a large dataset from scratch.

\subsection{Research Questions}

With the training techniques outlined for different model update
scenarios, we consider four research questions in our experiments.

\paragraph{RQ0: Can an attacker learn private information from model updates?}
Here we address the basic question of whether private data used to
update a model can be leaked in our adversarial setting and how. We
first answer this question by using differential score to \emph{find}
information about private sequences used in a model update.  We then
investigate the influence of other parameters of the system on the
differential score in more detail.

\paragraph{RQ1: How does masking private data with additional non-sensitive
data ($D_{\mathit{extra}}$) affect leakage?}
This is particularly important for the user deletion scenario, for
which we need to answer if it is possible to safely remove data of a
single user, or if such dataset changes need to be hidden among other
substantial changes. Concretely, we analyze whether including a large
enough additional dataset $D_{\mathit{extra}}$ in an update can
prevent leakage of information about the rest of the data used.
$D_{\mathit{extra}}$ can be any dataset which is either available
publicly or is non-sensitive from the point of view of the model
provider or users.

\paragraph{RQ2: How do retraining and continued training differ with respect to information leakage?}
In the continued training approach, the parameters of a previously
trained model $M_D$ are updated based only on new data $D' \setminus
D$. In contrast, in the retraining strategy parameters are updated
using all data in $D'$. The most recent updates to model parameters
depend only on new data in the continuing training case, whereas they
depend on the whole training data $D'$ when retraining a model from
scratch. We analyze the effect of this seemingly more pronounced
dependence.

\paragraph{RQ3: How is leakage affected by an adversary's background knowledge?}
Prior attacks on language models assume that the adversary has
background knowledge about the context in which a secret appears. We
analyze the effect of such knowledge for inferring private data from
model updates.

\subsection{Results with Canaries}
\label{sec:canaries}
We create a number of canary phrases---grammatically correct phrases
that do not appear in the original dataset---that serve as a proxy for
private data that the adversary is trying to extract. We consider
different word frequency characteristics to control the influence on
the used vocabulary.
Specifically, we fix the length of the canary phrase to $5$, choose a
valid phrase structure (e.g., Subject, Verb, Adverb, Compound Object),
and instantiate each placeholder with a token in a dataset
vocabulary. We create canaries in which frequencies of tokens are
{\em all low} (all tokens are from the least frequent quintile of words),
{\em mixed} (one token from each quintile),
{\em increasing from low to high}, and
{\em decreasing from high to low}.
For example, the \emph{mixed} phrase across all the datasets is ``NASA
used deadly carbon devices'', and the \emph{all low} phrase for PTB is
``nurses nervously trusted incompetent graduates''.
As the vocabularies differ between the different datasets, the
canaries are in general dataset-dependent.
We vary the amount of {\em private data}, $C$, by inserting a canary
phrase $s$ a number of times proportional to the number of tokens in
the training corpus:
\begin{asparaenum}
\item For PTB, we consider $k \in \{10,50,100\}$ canary insertions
  (corresponding to 1 canary token in 18K training tokens, 1 in 3.6K,
  and 1 in 1.8K).
\item For the Reddit dataset, we use $k \in \{5,50,500\}$
  (corresponding to 1 in 1M, 1 in 100K, 1 in 10K).
\item For the Wikitext-103 data, we use $k \in \{20, 100\}$
  (corresponding to 1 in 1M, 1 in 200K).
\end{asparaenum}

We train the model $M$ on $D$ and the model $M'$ on $D$ with $k$
copies of the canary $s$.
We then compute the differential rank of the canaries for different
values of $k$.

\begin{table*}[t]
  \caption{
  Differential score ($\dscore{}{}$) for different datasets, model architectures, canaries, and insertion frequencies.
  White cells represent a differential rank ($\drank$) of 0 (as approximated by beam search), and gray cells represent $\drank > 1000$.}
  \label{tbl:canary_results}
  \small
  \begin{tabular}{@{}lcccl@{\quad}cccl@{\ \ }cccl@{\quad}ccl@{}}
   \toprule
   Dataset
     & \multicolumn{3}{c}{Penn Treebank} &
     & \multicolumn{6}{c}{Reddit} &
     & \multicolumn{3}{c}{Wikitext-103}
     \\
   \cmidrule{2-4} \cmidrule{6-12} \cmidrule{14-15}
   Model Type (Perplexity)
     & \multicolumn{3}{c}{RNN (120.90)} &
     & \multicolumn{3}{c}{RNN (79.63)} &
     & \multicolumn{3}{c}{Transformer (69.29)} &
     & \multicolumn{2}{c}{RNN (48.59)}
     \\
   \cmidrule{2-4} \cmidrule{6-8} \cmidrule{10-12}\cmidrule{14-15}
   Canary Token Freq.
     & 1:18K & 1:3.6K & 1:1.8K &
     & 1:1M  & 1:100K & 1:10K  &
     & 1:1M  & 1:100K & 1:10K  &
     & 1:1M  & 1:200K
     \\
   \cmidrule(r){2-2} \cmidrule(lr){3-3} \cmidrule(l){4-4}
   \cmidrule(){6-6} \cmidrule(lr){7-7} \cmidrule(l){8-8}
   \cmidrule(){10-10} \cmidrule(lr){11-11} \cmidrule(l){12-12}
   \cmidrule(){14-14} \cmidrule(l){15-15}
   All Low
     & 3.40  & 3.94  & 3.97 &
     & 2.83  & 3.91  & 3.96 &
     & 3.22  & 3.97  & 3.99 &
     & \cellcolor{lightgray}1.39 & 3.81
     \\
   Low to High
     & 3.52  & 3.85  & 3.97 &
     & \cellcolor{lightgray}0.42 & 3.66 & 3.98 &
     & \cellcolor{lightgray}0.25 & 3.66 & 3.97 &
     & \cellcolor{lightgray}0.07 & 3.21
     \\
   Mixed
     & 3.02  & 3.61  & 3.90 &
     & \cellcolor{lightgray}0.23 & 3.04 & 3.92 &
     & \cellcolor{lightgray}0.39 & 3.25 & 3.96 &
     & \cellcolor{lightgray}0.25 & 3.02
     \\
   High to Low
     & \cellcolor{lightgray}1.96 & 2.83 & 3.46 &
     & \cellcolor{lightgray}0.74 & \cellcolor{lightgray}1.59 & 2.89 &
     & \cellcolor{lightgray}0.18 & \cellcolor{lightgray}1.87 & 3.10 &
     & \cellcolor{lightgray}0.08 & \cellcolor{lightgray}1.22
     \\
   \bottomrule
  \end{tabular}
\end{table*}

\begin{table*}[t]
  \caption{Differential Score ($\dscore{M}{M'}$) of the mixed frequency
    canary phrase for the Reddit (RNN) model using different update techniques.
    Model $M$ is trained on $D_{\mathit{orig}}$.
    For the \emph{Retraining} column, $M'$ is trained on $D_{\mathit{orig}} \cup D_{\mathit{extra}} \cup C$ starting from random initial parameters.
    For the \emph{Cont'd Training 1} column, $M'$ is trained on $D_{\mathit{extra}} \cup C$ starting from $M$.
    For the \emph{Cont'd Training 2} column, we first train a model $\tilde{M}$ on $D_{\mathit{extra} }\cup C$ starting from $M$, and then train
    model $M'$ from $\tilde{M}$ using additional public data $D'_{\mathit{extra}}$.
    A white cell background means that the differential rank $\drank$
    (as approximated by our beam search) of the phrase is 0, gray cell background means that $\drank$ is >1000.}
  \label{tbl:extra_results}
  \small
  \begin{tabular}{@{}lcccl@{\quad}cccccccl@{\quad}ccl@{}}
    \toprule
    ~
      & \multicolumn{4}{c}{Retraining} &
      & \multicolumn{3}{c}{Continued Training 1} &
      & \multicolumn{1}{c}{Continued Training 2}
      \\
    \cmidrule{2-5} \cmidrule{7-9} \cmidrule{11-11}
    ${|D_{\mathit{extra}}|} / {|D_{\mathit{orig}}|}$
      & $0\%$ & $20\%$ & $50\%$ & $100\%$ &
      & $20\%$ & $50\%$ & $100\%$ &
      & $100\%$
      \\
    \cmidrule(r){2-2} \cmidrule(lr){3-3} \cmidrule(lr){4-4} \cmidrule(lr){5-5} 
    \cmidrule(r){7-7} \cmidrule(lr){8-8} \cmidrule(lr){9-9}
    \cmidrule(lr){11-11} 
    1:1M
      & \cellcolor{lightgray}0.23 & \cellcolor{lightgray}0.224 & \cellcolor{lightgray}0.223 & \cellcolor{lightgray}0.229 &
      & \cellcolor{lightgray}0.52 & \cellcolor{lightgray}0.34  & \cellcolor{lightgray}0.46  &
      & \cellcolor{lightgray}0.01
      \\
    1:100K                     
      & 3.04 & 3.032 & 3.031 & 3.038 &
      & 3.56 & 3.25  & 3.27  &
      & \cellcolor{lightgray}0.26
      \\
    \bottomrule
  \end{tabular}
\end{table*}

\paragraph{RQ0: Can an attacker learn private information from model updates?}
We use our differential score based beam search
(Algorithm~\ref{alg:beam_search}) to extract canary phrases that
correspond to the change in training data between $M$ and $M'$. The
results of varying the number of inserted canaries are summarized
in~\autoref{tbl:canary_results}. We highlight the following findings:

\begin{asparaitem}
\item \textbf{For most combinations of $k$ and types of canaries, we
  successfully recover the canary.} This is indicated by the cells
  with white background, where the canary phrase has the {\em maximum
    differential score} among all token sequences found by our beam
  search, i.e., it ranks first.
\item The signal for extraction is strong even when the inserted
  canaries account for only {0.0001\%} of the tokens in the dataset.
  This is visible in the first row of \autoref{tbl:canary_results}
  where differential scores approach $4$ --- close to the upper
  bound of $5$ for $5$-token canaries.
\item Private phrases that occur more often in the training data are
  more exposed via a model update, as expected. This is visible in the
  monotonic growth of the differential score of canaries with the
  number of insertions.
\item Phrases composed of rare words are more easily extracted, as
  seen in the high differential score of canaries constructed from
  low-frequency tokens. In contrast, canaries with descending token
  frequencies tolerate much higher number of insertions before being
  exposed. This is expected, as our beam search is biased towards
  finding high-scoring prefixes.
\item Access to two model snapshots reveals substantially more than
 access to a single snapshot. For comparison, we successfully extract
 a 5-token canary inserted 1 in 200k times (i.e. inserting one token
 every 1M tokens) from two snapshots of an LSTM-based generative model
 without additional knowledge. In contrast, \citep[Section
 6.1]{secret-sharer} reports failing to extract the middle token of a
 5-token canary inserted 1 in 100k times from a similar LSTM-based
 model when given the first and last two words.
\end{asparaitem}

\paragraph{RQ1: Effect of amount of public vs. private data.}
In Table~\ref{tbl:extra_results} we vary the amount of {\em public data} by partitioning the dataset
$D$ into $D_{\mathit{orig}}\uplus D_{\mathit{extra}}$ such that the
latter is $20\%$, $50\%$, or $100\%$ of the size of $D_{\mathit{orig}}$
(the $0\%$ column is identical to Table~\ref{tbl:canary_results}).
%
%
The retraining column shows that
$\dscore{M}{M'}$ does not change significantly across the different
dataset splits.
That is, canaries can be extracted from the trained model even when
they are contained in a substantially larger dataset extension.
Hence, the amount of public data in the update does not
significantly affect the leakage of the private data.

\paragraph{RQ2: Effect of training type.}
We train a model $M$ on a dataset $D_{\mathit{orig}}$ to convergence,
and then continue training $M$ using $D_{\mathit{extra}}$ and the
canaries $C$, obtaining $M'$.
We compare the differential rank of the canaries on the models
obtained using continued training with that on the models retrained
from scratch (shown in the middle column of Table~\ref{tbl:extra_results}).
%
%
We observe that in all cases the differential score is higher for
continued training than for retraining.
As expected, the differential score of the canary phrase decreases as
additional extra data is used for fine-tuning.

\paragraph{RQ3: Effect of background knowledge.}
We evaluate the differential score of suffixes of a canary phrase $s$
assuming knowledge of a prefix.
For $i=1,\dots, n$ we take the prefix $t_1\dots t_{i-1}$ of the canary
phrase and compute the differential score $r$ of the token $t_{i}$
conditional on having read the prefix, i.e., $M'(t_{<i})(t_i) -
M(t_{<i})(t_i)$.
The relationship between $i$ and $r$ indicates how much knowledge
about $s$ is required to expose the remainder of the canary phrase.

\begin{figure}[t]
  \begin{adjustbox}{width=\columnwidth,center}
  \includegraphics{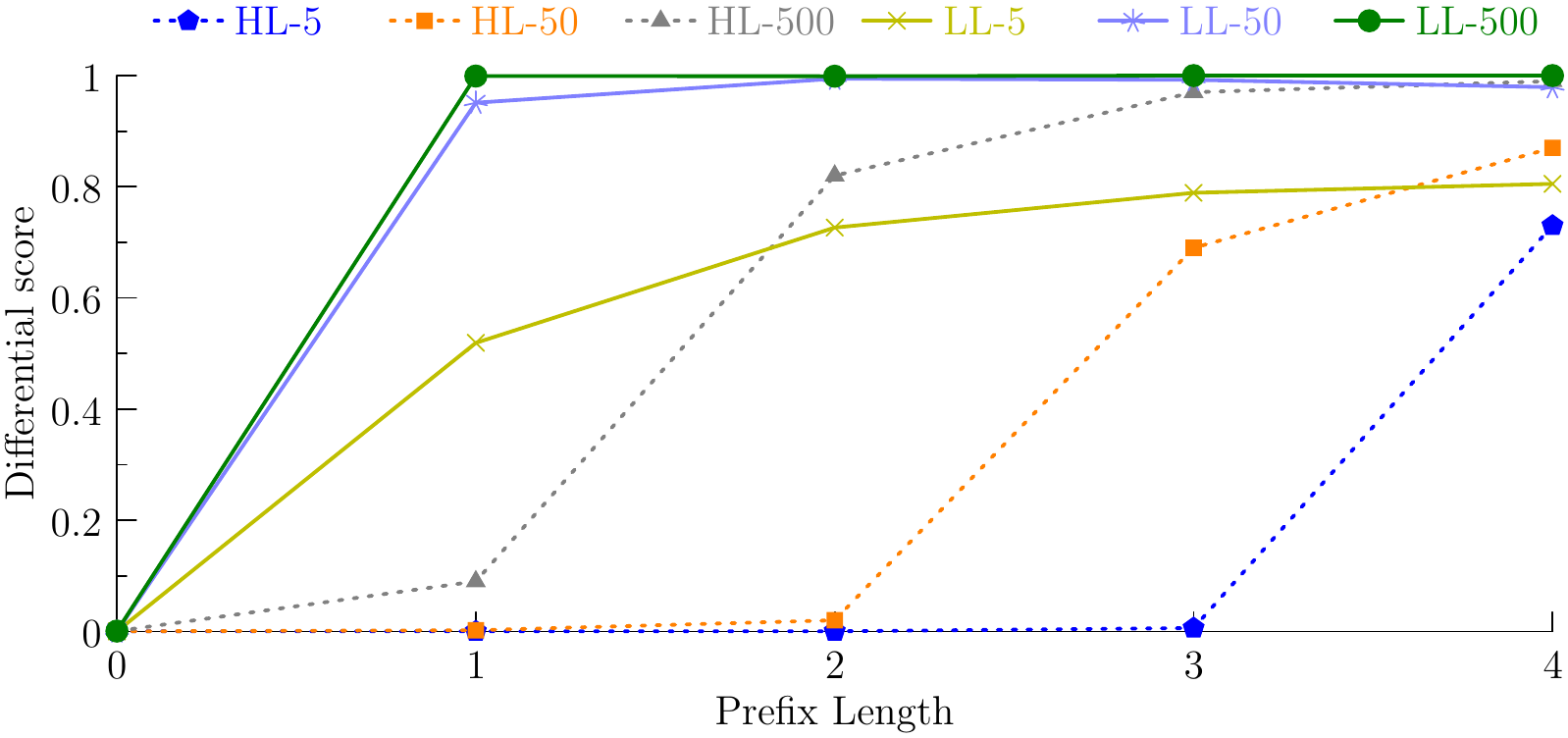}
  \end{adjustbox}
  \caption{Line graph showing that the differential score of canaries with all-low token frequencies rises faster with respect to prefix length compared to canaries with high-to-low token frequencies.}
  \caption{Differential score of tokens in canaries given a prefix for the Reddit dataset.
  LL-$k$ denotes $k$ canary insertions with all-low token frequencies (solid lines), and HL-$k$ denotes high-to-low token frequencies (dashed lines).}
  \label{fig:context_results}
\end{figure}

\autoref{fig:context_results} depicts the result of this analysis for
canaries with high-to-low and all-low token frequencies on the Reddit
dataset.
Our results show that, while the differential score of the first token
without context is close to 0, the score of subsequent tokens quickly
grows for all-low canaries, even with a low number of canary
insertions.
In contrast, more context is required before the
score of high-to-low canaries increases, as the model is less influenced by the
small number of additional occurrences of frequent tokens.

This suggests that, even in cases where we fail to extract the canary
without additional knowledge, an adversary can use the differential
rank to complete a partially known phrase, or confirm that a phrase
was used to update the model.

\subsection{Results with Real-world Data}
\label{sec:res_real_world}
We simulate real-world scenarios by sourcing training data from
real-world conversations on specific topics, and using it as a proxy
for private data included in the training data used in model updates.
The adversary's goal is to extract specific phrases occurring in the
proxy dataset, or phrases that do not occur literally but nonetheless
reveal the topic of conversations.

We mimic the data distribution shift by choosing conversations on
topics that are not dominant in the original dataset, so that we can
better judge whether phrases extracted using differential score are
on-topic and thus represent meaningful leakage of private information.
Specifically, we compare models trained only on data from the Reddit
dataset against models trained on data from the Reddit dataset plus
messages from one of two newsgroups from the 20 Newsgroups
dataset~\citep{Lang:ICML95}:
\begin{enumerate}[a)]
\item \texttt{rec.sport.hockey},~containing around 184K tokens,
   $\approx$1\% of the original training data; and
\item \texttt{talk.politics.mideast},~containing around 430K tokens,
   $\approx$2\% of the original training data.
\end{enumerate}

We train a model $M$ on the entire Reddit dataset and retrain $M'$
from scratch on the same dataset plus all messages from one of the two
newsgroups.
For both model architectures (RNNs and Transformer) described
in \autoref{sec:expsetup} and each newsgroup, we compute the sequences
with highest relative differential score.
Since the sequences returned by vanilla beam search typically share a
common prefix, we run a group beam search (see Section~\ref{sec:algo})
to get a more diverse sample.

\paragraph{RQ0: Can an attacker learn private information from model updates?}
Tables~\ref{tbl:trend_results} and~\ref{tbl:mideast_results} (in the
Appendix) display the highest-scoring sequences of length 4 in each
group of a $\drelscore{}{}$-based 5-group beam search.

\begin{table*}[ht]
  \caption{Top ranked phrases in group beam search for a model updated with \texttt{rec.sport.hockey}.
    For the layperson: Los Angeles Kings, Minnesota North Stars, and Toronto Maple Leaf are
    National Hockey League teams; Norm Green was the owner of the North Stars; an ice
    hockey game consists of three periods with overtime to break ties.
    Capitalization added for emphasis.}
 \label{tbl:trend_results}
 \small
  \begin{tabular}{@{\quad}lc@{\quad}r@{\quad}lc}
    \toprule
       \multicolumn{2}{c}{RNN} & &     
       \multicolumn{2}{c}{Transformer}\\[-5pt]
      Phrase & $\drelscore{}{}$ & &
      Phrase & $\drelscore{}{}$ \\
      \cmidrule(r){1-1} \cmidrule(lr){2-2} \cmidrule(r){4-4} \cmidrule(lr){5-5}
    \tt Angeles Kings prize pools         & 56.42 & &
    \tt Minnesota North Stars playoff     & 96.81 \\   
    \tt National Hockey League champions  & 53.68 & &
    \tt Arsenal Maple Leaf fans           & 71.88 \\
    \tt Norm 's advocate is               & 39.66 & &
    \tt Overtime no scoring chance        & 54.77 \\
    \tt Intention you lecture me          & 21.59 & &
    \tt Period 2 power play               & 47,85 \\
    \tt Covering yourself basically means & 21.41 & &
    \tt Penalty shot playoff results      & 42.63 \\
  \bottomrule
  \end{tabular}
\end{table*}

\textbf{The exposed sentences are on-topic} w.r.t.~the newsgroup included, e.g.,
the hockey theme dominates the top ranked sequences in
Table~\ref{tbl:trend_results}. This suggests that, information about
the private data used for the update is leaked.
It is noteworthy that these results are obtained assuming a weak
adversary that does not require either background knowledge
about the dataset distribution or about the information it tries to
extract.
In contrast, concurrent work on updates of image classification
models~\cite{DBLP:journals/corr/abs-1904-01067} requires knowledge
about the data distribution to train shadow models, while prior work
on single language models~\citep{secret-sharer} requires a known
prefix for extraction of a secret.

Given some background knowledge in the form of a long enough prefix of
a phrase occurring in the private data, we show that the complete
phrase can be extracted by a beam search directed by differential
score~(see Table~\ref{tbl:trends_prefixes}).

\begin{table*}[ht]
  \caption{Relative differential score of phrases found by beam search
    when retraining from scratch and continuing training from a
    previous model. The results are for RNN models trained on
    partitions of the Reddit dataset with
    $N = \texttt{talk.politics.mideast}$. Cells for which continued
    training yields a higher score than retraining
    appear in bold font. Capitalization added for emphasis.}
 \label{tbl:trend_continued}
 \newcommand{\HL}{\bf}
 \small
 \begin{tabular}{@{}lrrrrr@{\hspace{3ex}}rrrrr@{~~~~}} \toprule
   & \multicolumn{5}{c}{Retraining} & \multicolumn{5}{c}{Continued Training} \\[-12pt]
   \multicolumn{1}{c}{Phrase (\# of occurrences in $N$)}
          & \multicolumn{5}{c}{}
          & \multicolumn{5}{c}{}
          \\
   \cmidrule(lr){2-6} \cmidrule{7-11}
   \multicolumn{1}{r}{$|D_\mathit{extra}|/|D_\mathit{orig}|$}
          & 0\%      & 5\%    & 10\%   & 20\%   & 100\%
          & 0\%      & 5\%    & 10\%   & 20\%   & 100\% \\
   \multicolumn{1}{r}{Perplexity decrease}
          & 0.79  & 1.17  & 2.45  & 3.82  & 11.82
          & 73.97 & 18.45 & 10.29 & 6.08  & 8.28 \\
   \midrule
   {\tt Center for Policy Research} (93)
   & 99.77  & 101.38 & 97.11  & 98.65  & 91.53
   & \HL 276.98 & \HL 198.69 & \HL 150.56 & \HL 122.25 & \HL 117.54 \\
   {\tt Troops surrounded village after} (12)
   & 44.50  & 44.50  & 44.50  & 44.41  & 44.54
   & \HL 173.95 & \HL 47.38  & 19.48  & 7.81   & 35.56 \\
   {\tt Partition of northern Israel} (0)
   & 27.61  & 16.81  & 38.48  & 26.10  & 38.76
   & \HL 68.98  & 16.48  & 12.47  & 22.93  & 18.82 \\
   {\tt West Bank peace talks} (0)
   & 25.68  & 25.64  & 25.69  & 25.71  & 25.75
   & \HL 71.54  & 24.38  & \HL 28.60  & 16.91  & 4.62  \\
   {\tt Spiritual and political leaders} (0)
   & 25.23  & 25.98  & 17.04  & 24.21 & 23.47
   & \HL 126.92 & 14.91  & 10.00  & 3.44  & 11.05 \\
   {\tt Saudi troops surrounded village} (0)
   & 24.31  & 24.31  & 24.31  & 24.31  & 24.30
   & 5.05   & \HL 44.58  & 4.29   & 7.29   & \HL 63.84 \\
   {\tt Arab governments invaded Turkey} (0)
   & 22.59  & 22.62  & 22.80  & 22.78  & 22.80
   & \HL 24.01  & 15.58  & 7.08   & 18.12  & 11.90 \\
   {\tt Little resistance was offered} (12)
   & 22.24  & 22.09  & 25.12  & 22.34  & 25.59
   & \HL 215.16 & \HL 25.02 & 2.00  & 3.30  & 5.64 \\
   {\tt Buffer zone aimed at protecting} (0)
   & 4.00   & 4.47   & 5.30   & 5.25   & 5.69
   & \HL 57.29  & \HL 69.76  & \HL 18.92  & \HL 14.50  & \HL 22.25 \\
   {\tt Capital letters racial discrimination} (0)
   & 3.76   & 3.32   & 3.40   & 3.60   & 3.84
   & \HL 94.60  & \HL 52.74  & \HL 39.11  & \HL 11.22  & 3.45 \\
   \bottomrule
 \end{tabular}
\end{table*}

\paragraph{RQ1: Effect of amount of public vs. private data.}
We consider partitions of the Reddit dataset $D$ into
$D_{\mathit{orig}}$ and $D_{\mathit{extra}}$ of different relative
sizes.
For each partition, we train a model $M$ on $D_{\mathit{orig}}$ and a
model $M'$ on $D_{\mathit{orig}}\cup D_{\mathit{extra}} \cup N$, where
$N$ are all messages from \texttt{talk.politics.mideast}.
We observe the following:
\begin{asparaitem}
\item For all phrases, the proportion of public data ranging from $5$\% to
$100$\% used in the update does not significantly affect their
relative differential scores, which confirms our findings for
canaries.
\item The top two phrases resemble canaries in that they occur literally multiple
times in the update dataset, which explains their high scores. An
exception is \texttt{Little resistance was offered}, which appears
$12$ times in the dataset but still has low score.
Other phrases do not occur literally in newsgroup messages, but digest
recurrent discussions or contain $n$-grams that do occur.
\end{asparaitem}

\paragraph{RQ2: Effect of training type.}
We train a model $M$ on $D_{\mathit{orig}}$ to convergence, and then
continue training $M$ using $D_{\mathit{extra}} \cup N$ to produce a
model $M'$.
To understand the effect of the training type on information leakage,
we sample a set of representative phrases and compare their relative
differential scores w.r.t. $M$ and $M'$ against their scores
w.r.t. $M$ and a model trained on $D\cup N$ from scratch.

The results are shown in \autoref{tbl:trend_continued}, together with
the perplexity decrease after the model update.
Retrained models correspond to the \emph{data update} and \emph{data
deletion} scenarios and their perplexity drop is greater the more data
is used during retraining.
Continued training corresponds to the \emph{data specialization}
scenario. The perplexity drop in the updated model is greater the
larger is the proportion of newsgroup data used in the update, for
which the initial model is not specialized.

The last two rows in \autoref{tbl:trend_continued} correspond to
phrases found by group beam search in the continued training scenario,
but that have too low a score to be found when $M'$ is retrained from
scratch instead. The converse, i.e., phrases that have low score when
continuing training and high score when retraining, seems to occur
rarely and less consistently (e.g., \texttt{Saudi troops surrounded
village}).

For phrases that occur literally in the dataset, the results are in line with those for canaries (see
\autoref{tbl:extra_results}), with scores decreasing as more data is
used during the fine-tuning stage. For other phrases, the results are
not as clear-cut.
While fine-tuning a model exclusively on private data yields scores
that are significantly higher than when retraining a model from
scratch, this effect vanishes as more additional data is used; in some
cases continued training yields scores lower than when retraining a
model on the same data.

\paragraph{RQ3: Effect of background knowledge.}

An adversary wishing to extract information about the dataset used to
update a language model may direct a search using as
\emph{prompt} a known prefix from the dataset. We study how
long this prefix needs to be to recover the rest of phrase.

We consider a RNN model $M$ trained on the full Reddit dataset and a
model $M'$ trained on the union of the full Reddit dataset and all
messages of the \texttt{talk.politics.mideast} newsgroup.
We sample 4 phrases in newsgroup messages beginning with the name of a
Middle Eastern country and containing only tokens in the model
vocabulary.
We believe it is feasible for the adversary to guess these prefixes from the description of the newsgroup or the geopolitical
context.
For each phrase $s$ and $i=0,\dots, |s|-1$ we run a
$\drelscore{}{}$-based beam search for phrases of the same length with
constant beam width \num{10000} and 100 groups starting from $s_1\dots
s_{i}$.
Table~\ref{tbl:trends_prefixes} shows the rank of $s$ among the search results (or $\infty$ if absent).

\begin{table*}[t]
  \caption{Results of beam searches for different prefix lengths.
  A rank of 0 means that the search recovers the complete phrase.
  Due to the heuristic nature of the search the rank reported may be
  lower than the true rank of $s$.
  Conversely, a beam search may not encounter $s$ at all despite having
  lower rank than most phrases encountered.
  For instance, this occurs for \texttt{Turkey searched an American plane},
  where all but 7 search results with no prompt have higher rank (lower score).
  %
  }
  \label{tbl:trends_prefixes}
  \small
  \begin{tabular}{l@{~~~}cccccccc}
    \toprule
    & & & \multicolumn{6}{c}{Prefix length $i$} \\
    \cmidrule{4-9}
    \multicolumn{1}{c}{Phrase $s$} & \# of occurrences & $\drelscore{}{}(s)$ & 0 & 1 & 2 & 3 & 4 & 5 \\
    \cmidrule{1-9}
    \texttt{Turkey searched an American plane}    & 6 & 82.96 
      & $\infty$ & 1 & 1 & 0 & 0 & -- \\

    \texttt{Israel allows freedom of religion}    & 3 & 24.44 
      & $\infty$ & $\infty$ & 788 & 55 & 0 & -- \\

    \texttt{Iraq with an elected government}      & 2 & 23.75 
      & $\infty$ & $\infty$ & $\infty$ & 4 & 0 & -- \\   

    \texttt{Israel sealed off the occupied lands} & 2 & 6.48 
      & $\infty$ & $\infty$ & $\infty$ & $\infty$ & 3442 & 2 \\
    \bottomrule
  \end{tabular}
\end{table*}

We observe a correlation between the score of a phrase and the minimum
prefix sufficient to recover it.
However, a dip in the score of two consecutive tokens is much more
consequential: a common word like \texttt{the}, which has a similar
distribution in the original and private datasets, contributes little
to the score of a phrase and is unlikely to be picked up as a
candidate extension in a beam search.
Recovering from this requires additional heuristics or a more
expensive search, using wider beams or looking more than one token
ahead to better approximate the true rank of a phrase.

\section{Characterizing the Source of Leakage}
\label{sec:leaksource}
Prior work has primarily studied information leakage when an attacker
has only access to a single model snapshot.
Here, we first analyze how much our analysis gains from having access
to two model snapshots, and then consider the influence of common
causes of leakage in the single-model case.
The central ones are {\em overfitting}~\cite{YeomGFJ18} to the
training data, and {\em unintended memorization}~\cite{secret-sharer}
of data items that is independent of the distribution to be learned.

\paragraph{RQ4: How important is access to a second model snapshot?}
We want to analyze how much leakage of sensitive information is
increased when having access to two model snapshots $M_D$, $M_{D'}$ in
contrast to having only access to a single model $M_{D'}$.
This is a challenging analysis in a realistic setting, due to the size
of the data and the lack of an easily computable metric for
information leakage. Concretely, we want to show that the data we can
extract using the differential analysis of $M_D$ and $M_{D'}$ is
 (a) more likely to be part of $D'$ than of $D$,
 (b) not very common in $D'$,
 and
 (c) that (a) and (b) are more true for the results of the differential analysis
  than for the analysis of $M_{D'}$ alone.

We quantify how likely a given sentence is to be a part of a dataset
using a simpler, well-understood model of natural language data,
namely an $n$-gram model.
$n$-gram models define the probability of a token $t_{n+1}$ appearing
after a sequence of tokens $t_1 \ldots t_n$ as the number of times
$t_1 \ldots t_n t_{n+1}$ appeared in the dataset divided by the number
of times $t_1 \ldots t_n$ appeared.
%

In our experiments, we use the perplexity of $3$-gram models trained
on $D$ (resp. $N$) to capture how likely a given extracted sentence is
part of the dataset $D$ (resp. $N$).
We compare these perplexity values for sequences extracted using group
beam search from the models $M_D$ (resp. $M_{D'}$) and for sequences
extracted using our differential rank-based search, following the
setup of Section \ref{sec:res_real_world}.
Concretely, we used the entire Reddit comment data as dataset $D$, and the
messages $N$ from \texttt{talk.politics.mideast} as data update.
We are concerned with information an attacker can gain about the
contents of $N$.

\begin{figure*}[t]
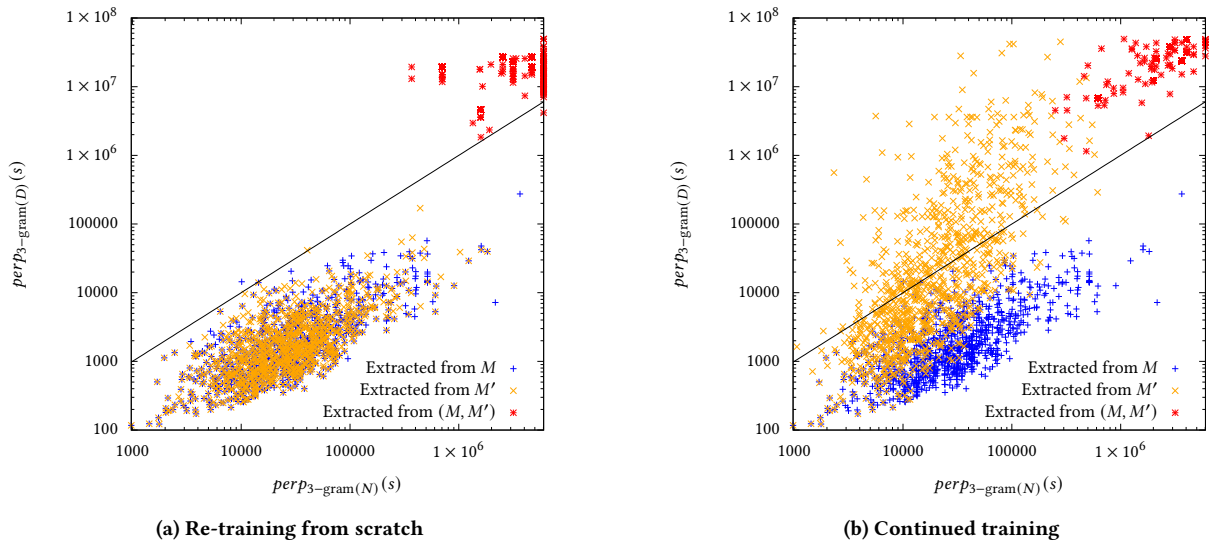

  \begin{subfigure}{.49\textwidth}
    \begin{adjustbox}{width=1.1\textwidth,center}
    \input{figures/plot_trainingtype_mixed_wrapper}
    \end{adjustbox}
    \caption{Re-training from scratch}
    \label{fig:public_vs_priv_retrain}
  \end{subfigure}
  \begin{subfigure}{.49\textwidth}
    \begin{adjustbox}{width=1.1\textwidth,center}
    \input{figures/plot_trainingtype_finetuned_wrapper}
    \end{adjustbox}
    \caption{Continued training}
    \label{fig:public_vs_priv}
  \end{subfigure}
  \caption{Sensitivity of extracted content. 
    \textcolor{blue}{$+$} depict sentences extracted from $M$,
    \textcolor{orange}{$\times$} from $M'$, and
    \textcolor{red}{$\ast$} from $(M,M')$ using Differential Score.
    Vertical axis depicts the perplexity w.r.t data $D$, horizontal axis depicts perplexity w.r.t data update $N$.
    Points above the diagonal are closer in distribution to the (private) data update $N$ than to the base data $D$.}
\end{figure*}

Figure~\ref{fig:public_vs_priv_retrain} shows the results of our
analysis when we train $M_{D'}$ on $D' = D \cup N$ from scratch.
Points above the main diagonal are closer in distribution to the
(private) data update $N$ than to the base data $D$.
This shows that our attack extracts sequences using differential score
(represented by red crosses) that are more likely to be part of $N$
than of $D$, and that these sequences differ substantially from the
sequences obtained by a single-model analysis.
In fact, the sequences obtained by single-model analysis for $M_D$ and
$M_{D'}$ show little significant difference.
Note that the perplexity values $\perpl_{3\text{-gram}(D)}$ are very
high for some of the extracted sentences, as they use combinations of
tokens that never appear in the original training dataset $D$.
Similarly, Figure~\ref{fig:public_vs_priv} shows the results of this
analysis on the scenario in which we obtain $M_{D'}$ by specializing
the model $M_{D}$ by continuing training on the dataset $N$.
While our differential analysis again captures sequences more likely
to be part of the updated data $N$ than of the original data $D$, the
single-model analysis now also shows some of this effect.

\paragraph{RQ5: Is leakage due to overfitting or intended memorization?}
All models are trained using an early-stopping criterion that halts
training when the model does not improve on a separate validation set.
This effectively rules out overfitting to the training data.
Additionally, model training employs regularization strategies such as
dropout to further encourage the trained models to generalize to
unseen data.

We refer to the model's ability to reproduce verbatim fragments of the
training data as \emph{memorization} and call it \emph{intended} if
this is necessary to serve its purpose of generating natural language
(e.g., a model needs to memorize the token pair ``United States'', as
it is an extremely common combination) and \emph{unintended}
otherwise.

In the experimental results in Table~\ref{tbl:trend_continued}, we
have included the number of times that the phrases with the highest
differential scores appear in the update dataset.
Since some of these phrases do not appear verbatim, we also measure
how close these phrases are to phrases in the original and update
datasets.
Table~\ref{tbl:near_matches} shows the Levenshtein distance of
extracted phrases from Table~\ref{tbl:trend_continued} to their
nearest neighbor in either dataset. Generally, we find closer matches
in the update dataset.
While ``Center for Policy Research'' is a clear case of intended
memorization, as the name appears many times in email signatures,
other phrases appear rarely or never, indicating that our analysis
extracts phrases that need not be memorized to serve its purpose.
This is further supported by the results in
Table~\ref{tbl:trends_prefixes}, where extraction of complete
sentences such as ``Israel allows freedom of religion'' occurring as
few as three times in the dataset is possible. Overall, this indicates
that intended memorization is unlikely to explain our results.

Unintended memorization may occur for infrequent phrases. However, it
cannot alone explain our results, as shown by our success in recovering
canaries when using a low-capacity model in a large-data regime (cf.
Wikitext-103 column in Table~\ref{tbl:canary_results}), for which the
effect of unintended memorization is less pronounced, and evidenced by
the large context needed to recover canaries from a single-model
analysis~\cite{secret-sharer}.
The most likely explanation remains that a differential analysis of
two model snapshots amplifies otherwise imperceptible differences in
the data used to train them, which would be hard to suppress without
hurting a model's performance.

\begin{table*}[ht]
  \caption{Quantifying near matches of extracted phrases from RNN models trained on the base Reddit dataset and updated with \texttt{talk.politics.mideast}. For each extracted phrase, we compare the Levenshtein distance to its nearest neighbor in the base and update datasets respectively.
  The updated dataset contains closer matches for all phrases except {\tt west bank peace talks} and {\tt capital letters racial discrimination}, for which there are equally close matches in both datasets.}
 \label{tbl:near_matches}
 \newcommand{\HL}{\bf}
 \small
 \begin{tabular}{@{}lrr@{\hspace{3ex}}rr@{~~~~}} \toprule
   Extracted phrase & \multicolumn{2}{c}{\texttt{talk.politics.mideast}} & \multicolumn{2}{c}{Reddit} \\
   \cmidrule(r){1-1} \cmidrule(lr){2-3} \cmidrule(l){4-5}
   {\tt center for policy research}
   & {\tt center for policy research} & 0
   & {\tt center for instant research} & 1 \\
   {\tt troops surrounded village after}
   & {\tt troops surrounded village after} & 0
   & {\tt from the village after} & 2 \\
   {\tt partition of northern israel}
   & {\tt shelling of northern israel} & 1
   & {\tt annexation of northern greece} & 2 \\
   {\tt west bank peace talks}
   & {\tt . no peace talks} & 2
   & {\tt : stated peace talks} & 2 \\
   {\tt spiritual and political leaders}
   & {\tt spiritual and political evolutions} & 1
   & {\tt , and like leaders} & 2 \\
   {\tt saudi troops surrounded village}
   & {\tt our troops surrounded village} & 1
   & {\tt " hometown " village} & 3 \\
   {\tt arab governments invaded turkey}
   & {\tt arab governments are not} & 2
   & {\tt ! or wrap turkey} & 3 \\
   {\tt little resistance was offered}
   & {\tt little resistance was offered} & 0
   & {\tt , i was offered } & 2 \\
   {\tt buffer zone aimed at protecting }
   & {\tt " aimed at protecting} & 2
   & {\tt 's aimed at a} & 3 \\
   {\tt capital letters racial discrimination} 
   & {\tt \% of racial discrimination} & 2 
   & {\tt allegory for racial discrimination} & 2  \\
   \bottomrule
 \end{tabular}
\end{table*}

\section{Mitigations}
\label{sec:mitigations}
In this section, we discuss and analyze three strategies to mitigate
information leakage in model updates:
\begin{inparaenum}
\item Differential Privacy,
\item continued training with public data, and
\item truncating the output of the updated model.
\end{inparaenum}

\subsection{Mitigation: Differential Privacy}

Differential privacy (DP)~\citep{privacybook} provides strong
guarantees on the amount of information leaked by a released output.
Given a computation over records it guarantees a bound on the effect
that any input record can have on the output.
Formally, $F$ is a $(\epsilon, \delta)$-differentially-private
computation if for any datasets $D$ and $D'$ that differ in one record
and for any subset $O$ of $F$'s range we have
\begin{displaymath}
  \Pr(F(D) \in O) \le \exp(\eps) \cdot \Pr(F(D') \in O) + \delta \,.
\end{displaymath}
Differential privacy is a natural candidate for defending against
membership-like inferences about data.
The exact application of differential privacy for protecting the
information in the model update depends on what one wishes to protect
w.r.t. the new data: individual sentences in the new data or all
information present in the update.
For the former, sequence-level privacy can suffice while for the
latter group DP can serve as a mitigation technique where the size of
the group is proportional to the number of sequences in the update.
Recall that an $\epsilon$-DP algorithm $F$ is
$k\epsilon$-differentially private for groups of size
$k$~\citep{privacybook}.

Differential privacy can be achieved in gradient-based optimization
computations~\citep{abadi,6736861,Bassily:2014:PER:2706700.2707412} by
clipping the gradient of every record in a batch according to some
bound $L$, then adding noise proportional to $L$ to the sum of the
clipped gradients, averaging over the batch size and using this noisy
average gradient update during backpropagation.

We evaluate the extent to which DP mitigates attacks considered in
this paper by training models on the Penn Treebank (PTB) dataset with
canaries with sequence-level differential privacy.
We train DP models using the TensorFlow Privacy
library~\citep{dptflink} for two sets of $(\epsilon,\delta)$
parameters, $(5,\num{1e-5})$ and $(111,\num{1e-5})$, for two datasets:
PTB and PTB with 50 insertions of the all-low-frequency canary.
We rely on~\citep{dptflink} to train models with differentially
private stochastic gradient descent using a Gaussian noise mechanism
and to compute the overall privacy loss of the training phase.
As expected, the performance of models trained with DP degrades, in
our case from $\approx$23\% accuracy in predicting the next token on
the validation dataset to 11.89\% and 13.34\% for $\epsilon$ values of
5 and 111, respectively.

While the beam search with the parameters of
Section~\ref{sec:canaries} no longer returns the canary phrase for the
DP-trained models, we note that the models have degraded so far that
they are essentially only predicting the most common words from each
class (e.g., ``is'' when a verb is required) and thus, the result is
unsurprising.
We note that the guarantees of sequence-level DP formally do not apply
for the case where canary phrases are inserted as multiple sequences,
and that $\epsilon$ values for our models are high.
However, the $\epsilon$-analysis is an upper bound and similar
observations about the effectiveness of training with DP with high
$\epsilon$ were reported by \citet{secret-sharer}.

We further investigate the effect of DP training on the differential
rank of a canary phrase that was inserted 50 times.
Instead of using our beam search method to approximate the
differential rank, we fully explore the space of subsequences of
length two, and find that the $\drank$ for the two-token prefix of our
canary phrase dropped from 0 to \num{9458399} and \num{849685} for the
models with $\eps=5$ and $\eps=111$ respectively.  In addition, we
compare the differential score of the whole phrase and observe that it
drops from 3.94 for the original model to $\num{4.5e-4}$ and
$\num{2.1e-3}$ for models with $\eps=5$ and $\eps=111$, respectively.
Though our experiment results validate that DP can mitigate the
particular attack method considered in this paper for canary phrases,
the model degradation is significant.
In addition, the computational overhead of per-sequence gradient
clipping required by~\citep{dptflink} is substantial, making it
unsuitable for training high-capacity neural language models on large
datasets.

\begin{figure*}[ht]
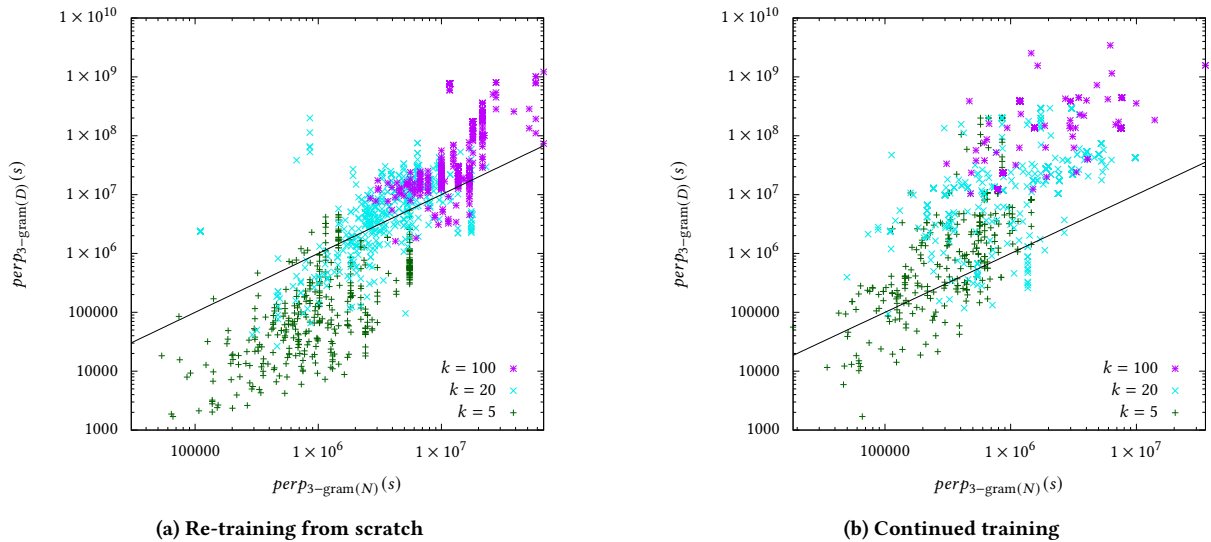

\begin{subfigure}{.49\textwidth}
  \begin{adjustbox}{width=1.1\textwidth,center}
  \input{figures/plot_mitigation_mixed_wrapper}
  \end{adjustbox}
  \caption{Re-training from scratch}
  \label{fig:topk-mixed}
\end{subfigure}
\begin{subfigure}{.49\textwidth}
  \begin{adjustbox}{width=1.1\textwidth,center}
  \input{figures/plot_mitigation_finetuned_wrapper}
  \end{adjustbox}
  \caption{Continued training}
  \label{fig:topk-finetuned}
\end{subfigure}
\caption{Sentences extracted from $(M,M')$ using Differential Score when the adversary only receives the top $k$ tokens from the updated model $M'$ for each query. The axes have the same meaning as in Figures~\ref{fig:public_vs_priv_retrain} and~\ref{fig:public_vs_priv}.}
\label{fig:topk}
\end{figure*}

\subsection{Mitigation: Two-stage Continued Training}

We also consider a possible mitigation strategy where we perform
continued training in two stages. For this, we split the dataset into
three equal parts $D_{\mathit{orig}}$, $D_{\mathit{extra}}$ and
$D_{\mathit{extra}}'$.
We proceed as in the continued training setting in RQ2, but add a
final step in which we train on another dataset after training on the
canaries.
This resembles a setting where an attacker does not have access to two
consecutive snapshots.
The rightmost column of Table~\ref{tbl:extra_results},
shows that the differential score of the canary phrase drops
substantially after the second training stage.
Thus, two or multi-stage continued training, where only the last
trained model is released, might be a path toward mitigating
leakage of private data.

\subsection{Mitigation: Truncating Output}
\label{sec:truncate}

Finally, we analyze the effect of truncating the output of the updated
model for each query.
Specifically, the adversary still has full access to the original
model $M$ but only receives the top $k$ tokens from the updated model
$M'$.
This is a slight weakening of our adversary model, but is realizable
for some applications.
For example, in the \emph{Data Specialization} scenario, the adversary
may have full access to the public base model, but can only access the
specialized model via an API that truncates the results for each
query.
In the \emph{Data Update} scenario, even if models are deployed to
client devices, it may be possible to enforce this by running the
model in a Trusted Execution Environment (TEE), such as Intel
SGX~\cite{SGX} or ARM TrustZone~\cite{TrustZone} on the client device.

To evaluate the impact of this mitigation, we repeat the experiment
described in Section~\ref{sec:leaksource} and plot only the sentences
extracted using differential score (i.e., the `Snapshot attack') for
different values of $k$.
To facilitate comparison, we use the same beam width as in
Figures~\ref{fig:public_vs_priv_retrain} and~\ref{fig:public_vs_priv}.
As shown in Figure~\ref{fig:topk}, decreasing the value of $k$ brings
the extracted sequences closer to the main diagonal, where they have
similar likelihood of being drawn from either dataset.
Similarly to Figures~\ref{fig:public_vs_priv_retrain}
and~\ref{fig:public_vs_priv}, we also observe a difference between
re-training from scratch and continued training; for the same value of
$k$, the sentences extracted after continued training are more likely
to be private than those extracted after the model is re-trained from
scratch.
Additionally, if the adversary only has access to the top $k$ outputs
of the original model $M$, this would further reduce the leakage.
In applications where this mitigation is realizable, returning only
the top $k$ outputs can thus reduce leakage without decreasing the
utility of the provided outputs.

\section{Related Work}
\label{sec:related}
Several works have shown that machine learning
models can leak information about  training data and proposed defenses for them.

\paragraph{Membership inference attacks.}
\citet{DBLP:conf/sp/ShokriSSS17} show
that one can identify whether a record belongs to the training dataset
of a classification model given black-box access to the model and
shadow models trained on data from a similar
distribution. \citet{Salem:NDSS19} demonstrate that similar attacks
are effective under weaker adversary models.
 It would be interesting to study how membership inference based
on differential score compares to other techniques~\cite{Chen:2020}.

\citet{natural-auditor} also study sequence-to-sequence language
models and show how a user can check if their data has been used for
training. In their setting, an auditor needs an auxiliary dataset to
train shadow models with the same algorithm as the target model and
queries the target model for predictions on a sample of the user's
data.
The auxiliary dataset does not need to be drawn from the same
distribution as the original training data (unlike
\cite{DBLP:conf/sp/ShokriSSS17}) and the auditor only observes a list
of several top-ranked tokens.
In contrast, our approach requires \emph{no} auxiliary dataset, but
assumes access to the probability distributions over all tokens from
two different model snapshots.
From this, we are able to recover full sequences from the differences
in training data rather than binary information about data presence.
Like them, we find that sequences with infrequent tokens provide a
stronger signal to the adversary/auditor.

\paragraph{Reconstruction attacks.}~These attacks
abuse a model to recover specific training points \citep{DBLP:journals/corr/abs-1904-01067}. The attacks we
present are a form of reconstruction attacks against an updated model:
we recover data points in the dataset used for the update given the
original model as auxiliary information.

\citet{secret-sharer} is closest to our work, as it also
considers information leakage of language models. The authors assess
the risk of (unintended) memorization of rare sequences in the
training data.
They show that canaries inserted into training data can be retrieved
from a character-level language model. The key differences to our
approach are that
\begin{inparaenum}[1)]
\item we consider a different attack scenario where \emph{an adversary has
  access to two snapshots of a model}, and
\item our canaries follow the distribution of the data whereas \citet{secret-sharer} add
  a random sequence of numbers in a fixed context into a dataset of
  financial news articles (e.g., ``The random number is ...''), where
  such phrases are rare. We instead are able to extract canaries {\em
  without any context}, even when the canary token frequency in the
  training dataset is as low as one in a million.
\end{inparaenum}

\citet{DBLP:journals/corr/abs-1904-01067} consider reconstruction of
training data that was used to update a model. While their goal is
similar to ours, their adversarial model and setup differ:
\begin{inparaenum}[1)]
\item similar to~\citet{natural-auditor} and \citet{DBLP:conf/sp/ShokriSSS17},
  their attacker uses shadow models trained on auxiliary data drawn
  from the same distribution as the target training dataset, while in
  our setting the attacker has no prior knowledge of this distribution
  and does not need auxiliary data;
\item the updated model is obtained by fine-tuning the target model
  with additional data rather than re-training it from scratch on the
  changed dataset;
\item the focus is on classification models and not on (generative)
  language models.
\end{inparaenum}

Information leakage from updates has also been considered for
searchable encryption: an attacker who has control over data in an
update to an encrypted database can learn information about its
content and previous encrypted searches on
it~\citep{Cash:2015:LAA:2810103.2813700}.

\paragraph{Model inversion attacks.}~\citet{Fredrikson:2014,Fredrikson:2015}
repurpose a model to work \emph{backwards}, inferring unknown
attributes of individuals given known attributes and a target
prediction. Individuals need not be present in the training data, and
results are aggregate statistics rather than information about
specific training points.

\paragraph{Differential Privacy.}
In terms of defenses, \citet{DBLP:conf/iclr/McMahanRT018} study how to
train LSTM models with DP guarantees at a user-level. They investigate
utility and privacy trade-offs of the trained models depending on a
range of parameters (e.g., clipping bound and batch size).
\citet{secret-sharer} show that DP protects against
leakage of canaries in character-level models,
while~\citet{natural-auditor} show that an audit as described above
fails when training language models with user-level DP using the
techniques of \cite{DBLP:conf/iclr/McMahanRT018}.
Pan-privacy~\citep{innovations:DworkNPRY10}, on the other hand,
studies the problem of maintaining differential privacy when an
attacker observes snapshots of the internal state of a DP algorithm
between updates.

\paragraph{Deletion of Data.}
Techniques to update models to delete training data points can be
broadly classified into \emph{exact} and \emph{approximate} deletion.
\citet{Ginart:NIPS2019} define
\emph{exact deletion} of a training point from a model as a stochastic
operation returning the same distribution as re-training from scratch
without that point, and develop deletion algorithms for $k$-means
clustering with low amortized cost.
\citet{bourtoule2019machine} propose an exact deletion methodology that
aggregates models trained on disjoint data shards, trading storage for
computation such that only shards that contain deleted points need to
be retrained.
Exact deletion is equivalent to retraining from scratch, hence,
publishing model snapshots before and after deletion matches our
adversarial model and our results apply.

Contemporary approximate deletion
methods~\citep{guo2019certified,Golatkar_2020_CVPR} yield models that
are only statistically indistinguishable from a model retrained from
scratch. These methods stochastically update model parameters based on
estimates of the influence of the data to be deleted and achieve
relaxations of differential privacy.
%
It would be interesting to study how susceptible to snapshot attacks
are models obtained by approximate deletion.


\section{Conclusion}
\label{sec:conclusion}
We presented a first systematic study of the privacy implications of
releasing snapshots of a language model trained on overlapping data.
Our results show that updates pose a threat which needs to
be considered in the lifecycle of machine learning applications.
We encourage the research community to work towards quantifying and
reducing unintended information leakage caused by model updates, and
hope to make practitioners aware of the privacy implications of
deploying and updating high-capacity language models.

\begin{acks}
We thank Doug Orr and Nicolas Papernot for helpful discussions and the
anonymous reviewers for their valuable comments.
\end{acks}

\begin{table*}[b]
  \caption{Top ranked phrases in a group beam search for a model updated with \texttt{talk.politics.mideast}.
     Center for Policy Research is a prolific newsgroup poster; many of the posts around the time the
     20 Newsgroups dataset~\citep{Lang:ICML95} was collected discuss tensions between Turkey and Armenia.}
  \label{tbl:mideast_results}
  \small
  \begin{tabular}{@{\quad}lc@{\quad}r@{\quad}lc}
  \toprule
     \multicolumn{2}{c}{RNN} & &
     \multicolumn{2}{c}{Transformer}\\[-5pt]
    Phrase & $\drelscore{}{}$ & &
    Phrase & $\drelscore{}{}$ \\
    \cmidrule(r){1-1} \cmidrule(lr){2-2} \cmidrule(r){4-4} \cmidrule(lr){5-5}
    \tt Turkey searched first aid         & 31.32 & &
    \tt Center for Policy Research        & 200.27 \\
    \tt Doll flies lay scattered          & 22.79 & &
    \tt Escaped of course ...             & 95.18 \\
    \tt Arab governments invaded Turkey   & 20.20 & &
    \tt Holocaust \%UNK\% museum museum   & 88.20 \\
    \tt Lawsuit offers crime rates        & 18.35 & &
    \tt Troops surrounded village after   & 79.35 \\
    \tt Sanity boosters health care       & 11.17 & &
    \tt Turkey searched neither Arab      & 37.69 \\
  \bottomrule
  \end{tabular}
\end{table*}

\balance

\bibliography{bibliography}


\begin{thebibliography}{00}


\ifx \showCODEN    \undefined \def \showCODEN     #1{\unskip}     \fi
\ifx \showDOI      \undefined \def \showDOI       #1{#1}\fi
\ifx \showISBNx    \undefined \def \showISBNx     #1{\unskip}     \fi
\ifx \showISBNxiii \undefined \def \showISBNxiii  #1{\unskip}     \fi
\ifx \showISSN     \undefined \def \showISSN      #1{\unskip}     \fi
\ifx \showLCCN     \undefined \def \showLCCN      #1{\unskip}     \fi
\ifx \shownote     \undefined \def \shownote      #1{#1}          \fi
\ifx \showarticletitle \undefined \def \showarticletitle #1{#1}   \fi
\ifx \showURL      \undefined \def \showURL       {\relax}        \fi
\providecommand\bibfield[2]{#2}
\providecommand\bibinfo[2]{#2}
\providecommand\natexlab[1]{#1}
\providecommand\showeprint[2][]{arXiv:#2}

\bibitem[\protect\citeauthoryear{Abadi, Chu, Goodfellow, McMahan, Mironov,
  Talwar, and Zhang}{Abadi et~al\mbox{.}}{2016}]%
        {abadi}
\bibfield{author}{\bibinfo{person}{Martin Abadi}, \bibinfo{person}{Andy Chu},
  \bibinfo{person}{Ian Goodfellow}, \bibinfo{person}{H.~Brendan McMahan},
  \bibinfo{person}{Ilya Mironov}, \bibinfo{person}{Kunal Talwar}, {and}
  \bibinfo{person}{Li Zhang}.} \bibinfo{year}{2016}\natexlab{}.
\newblock \showarticletitle{Deep Learning with Differential Privacy}. In
  \bibinfo{booktitle}{{\em 23rd ACM SIGSAC Conference on Computer and
  Communications Security, CCS 2016}}. \bibinfo{publisher}{ACM},
  \bibinfo{pages}{308--318}.
\newblock


\bibitem[\protect\citeauthoryear{Andrew, Chien, and Papernot}{Andrew
  et~al\mbox{.}}{2020}]%
        {dptflink}
\bibfield{author}{\bibinfo{person}{Galen Andrew}, \bibinfo{person}{Steve
  Chien}, {and} \bibinfo{person}{Nicolas Papernot}.}
  \bibinfo{year}{2020}\natexlab{}.
\newblock \bibinfo{title}{{TensorFlow Privacy}}.
\newblock \bibinfo{howpublished}{\url{https://github.com/tensorflow/privacy}}.
   (\bibinfo{year}{2020}).
\newblock


\bibitem[\protect\citeauthoryear{Arm}{Arm}{2020}]%
        {TrustZone}
\bibfield{author}{\bibinfo{person}{Arm}.} \bibinfo{year}{2020}\natexlab{}.
\newblock \bibinfo{title}{{TrustZone Technology}}.
\newblock   (\bibinfo{year}{2020}).
\newblock
\showURL{%
\url{https://developer.arm.com/ip-products/security-ip/trustzone}}


\bibitem[\protect\citeauthoryear{Bassily, Smith, and Thakurta}{Bassily
  et~al\mbox{.}}{2014}]%
        {Bassily:2014:PER:2706700.2707412}
\bibfield{author}{\bibinfo{person}{Raef Bassily}, \bibinfo{person}{Adam Smith},
  {and} \bibinfo{person}{Abhradeep Thakurta}.} \bibinfo{year}{2014}\natexlab{}.
\newblock \showarticletitle{Private Empirical Risk Minimization: Efficient
  Algorithms and Tight Error Bounds}. In \bibinfo{booktitle}{{\em 55th IEEE
  Annual Symposium on Foundations of Computer Science, FOCS 2014}}.
  \bibinfo{publisher}{IEEE Computer Society}, \bibinfo{pages}{464--473}.
\newblock


\bibitem[\protect\citeauthoryear{Bourtoule, Chandrasekaran, Choquette-Choo,
  Jia, Travers, Zhang, Lie, and Papernot}{Bourtoule et~al\mbox{.}}{2021}]%
        {bourtoule2019machine}
\bibfield{author}{\bibinfo{person}{Lucas Bourtoule}, \bibinfo{person}{Varun
  Chandrasekaran}, \bibinfo{person}{Christopher Choquette-Choo},
  \bibinfo{person}{Hengrui Jia}, \bibinfo{person}{Adelin Travers},
  \bibinfo{person}{Baiwu Zhang}, \bibinfo{person}{David Lie}, {and}
  \bibinfo{person}{Nicolas Papernot}.} \bibinfo{year}{2021}\natexlab{}.
\newblock \showarticletitle{Machine Unlearning}. In \bibinfo{booktitle}{{\em
  42nd IEEE Symposium on Security and Privacy, S\&P 2021}}.
  \bibinfo{publisher}{IEEE Computer Society}.
\newblock
\newblock
\shownote{To appear.}


\bibitem[\protect\citeauthoryear{Carlini, Liu, Erlingsson, Kos, and
  Song}{Carlini et~al\mbox{.}}{2019}]%
        {secret-sharer}
\bibfield{author}{\bibinfo{person}{Nicholas Carlini}, \bibinfo{person}{Chang
  Liu}, \bibinfo{person}{{\'{U}}lfar Erlingsson}, \bibinfo{person}{Jernej Kos},
  {and} \bibinfo{person}{Dawn Song}.} \bibinfo{year}{2019}\natexlab{}.
\newblock \showarticletitle{The Secret Sharer: Evaluating and Testing
  Unintended Memorization in Neural Networks}. In \bibinfo{booktitle}{{\em 28th
  {USENIX} Security Symposium}}. \bibinfo{publisher}{{USENIX} Association},
  \bibinfo{pages}{267--284}.
\newblock


\bibitem[\protect\citeauthoryear{Cash, Grubbs, Perry, and Ristenpart}{Cash
  et~al\mbox{.}}{2015}]%
        {Cash:2015:LAA:2810103.2813700}
\bibfield{author}{\bibinfo{person}{David Cash}, \bibinfo{person}{Paul Grubbs},
  \bibinfo{person}{Jason Perry}, {and} \bibinfo{person}{Thomas Ristenpart}.}
  \bibinfo{year}{2015}\natexlab{}.
\newblock \showarticletitle{Leakage-Abuse Attacks Against Searchable
  Encryption}. In \bibinfo{booktitle}{{\em 22nd ACM SIGSAC Conference on
  Computer and Communications Security, CCS 2015}}. \bibinfo{publisher}{ACM},
  \bibinfo{pages}{668--679}.
\newblock


\bibitem[\protect\citeauthoryear{Chen, Zhang, Wang, Backes, Humbert, and
  Zhang}{Chen et~al\mbox{.}}{2020}]%
        {Chen:2020}
\bibfield{author}{\bibinfo{person}{Min Chen}, \bibinfo{person}{Zhikun Zhang},
  \bibinfo{person}{Tianhao Wang}, \bibinfo{person}{Michael Backes},
  \bibinfo{person}{Mathias Humbert}, {and} \bibinfo{person}{Yang Zhang}.}
  \bibinfo{year}{2020}\natexlab{}.
\newblock \bibinfo{title}{When Machine Unlearning Jeopardizes Privacy}.
\newblock   (\bibinfo{year}{2020}).
\newblock
\showeprint[arxiv]{cs.CR/2005.02205}


\bibitem[\protect\citeauthoryear{Devlin, Chang, Lee, and Toutanova}{Devlin
  et~al\mbox{.}}{2019}]%
        {devlin2018bert}
\bibfield{author}{\bibinfo{person}{Jacob Devlin}, \bibinfo{person}{Ming-Wei
  Chang}, \bibinfo{person}{Kenton Lee}, {and} \bibinfo{person}{Kristina
  Toutanova}.} \bibinfo{year}{2019}\natexlab{}.
\newblock \showarticletitle{{BERT}: Pre-training of Deep Bidirectional
  Transformers for Language Understanding}. In \bibinfo{booktitle}{{\em 2019
  Conference of the North American Chapter of the Association for Computational
  Linguistics: Human Language Technologies, NAACL-HLT 2019}},
  Vol.~\bibinfo{volume}{1}. \bibinfo{publisher}{Association for Computational
  Linguistics}, \bibinfo{pages}{380--385}.
\newblock


\bibitem[\protect\citeauthoryear{Dwork, Naor, Pitassi, Rothblum, and
  Yekhanin}{Dwork et~al\mbox{.}}{2010}]%
        {innovations:DworkNPRY10}
\bibfield{author}{\bibinfo{person}{Cynthia Dwork}, \bibinfo{person}{Moni Naor},
  \bibinfo{person}{Toniann Pitassi}, \bibinfo{person}{Guy~N. Rothblum}, {and}
  \bibinfo{person}{Sergey Yekhanin}.} \bibinfo{year}{2010}\natexlab{}.
\newblock \showarticletitle{Pan-Private Streaming Algorithms}. In
  \bibinfo{booktitle}{{\em Innovations in Computer Science, ICS 2010}}.
  \bibinfo{publisher}{Tsinghua University Press}, \bibinfo{pages}{66--80}.
\newblock


\bibitem[\protect\citeauthoryear{Dwork and Roth}{Dwork and Roth}{2014}]%
        {privacybook}
\bibfield{author}{\bibinfo{person}{Cynthia Dwork} {and} \bibinfo{person}{Aaron
  Roth}.} \bibinfo{year}{2014}\natexlab{}.
\newblock \showarticletitle{The Algorithmic Foundations of Differential
  Privacy}.
\newblock \bibinfo{journal}{{\em Foundations and Trends in Theoretical Computer
  Science\/}} \bibinfo{volume}{9}, \bibinfo{number}{3-4}
  (\bibinfo{year}{2014}), \bibinfo{pages}{211--407}.
\newblock


\bibitem[\protect\citeauthoryear{Fredrikson, Jha, and Ristenpart}{Fredrikson
  et~al\mbox{.}}{2015}]%
        {Fredrikson:2015}
\bibfield{author}{\bibinfo{person}{Matt Fredrikson}, \bibinfo{person}{Somesh
  Jha}, {and} \bibinfo{person}{Thomas Ristenpart}.}
  \bibinfo{year}{2015}\natexlab{}.
\newblock \showarticletitle{Model Inversion Attacks that Exploit Confidence
  Information and Basic Countermeasures}. In \bibinfo{booktitle}{{\em 22nd
  {ACM} {SIGSAC} Conference on Computer and Communications Security, CCS
  2015}}. \bibinfo{publisher}{{ACM}}, \bibinfo{pages}{1322--1333}.
\newblock


\bibitem[\protect\citeauthoryear{Fredrikson, Lantz, Jha, Lin, Page, and
  Ristenpart}{Fredrikson et~al\mbox{.}}{2014}]%
        {Fredrikson:2014}
\bibfield{author}{\bibinfo{person}{Matthew Fredrikson}, \bibinfo{person}{Eric
  Lantz}, \bibinfo{person}{Somesh Jha}, \bibinfo{person}{Simon~M. Lin},
  \bibinfo{person}{David Page}, {and} \bibinfo{person}{Thomas Ristenpart}.}
  \bibinfo{year}{2014}\natexlab{}.
\newblock \showarticletitle{Privacy in Pharmacogenetics: An End-to-End Case
  Study of Personalized {W}arfarin Dosing}. In \bibinfo{booktitle}{{\em 23rd
  {USENIX} Security Symposium}}. \bibinfo{publisher}{{USENIX} Association},
  \bibinfo{pages}{17--32}.
\newblock


\bibitem[\protect\citeauthoryear{Ginart, Guan, Valiant, and Zou}{Ginart
  et~al\mbox{.}}{2019}]%
        {Ginart:NIPS2019}
\bibfield{author}{\bibinfo{person}{Antonio Ginart}, \bibinfo{person}{Melody
  Guan}, \bibinfo{person}{Gregory Valiant}, {and} \bibinfo{person}{James~Y
  Zou}.} \bibinfo{year}{2019}\natexlab{}.
\newblock \showarticletitle{Making {AI} Forget You: Data Deletion in Machine
  Learning}. In \bibinfo{booktitle}{{\em Advances in Neural Information
  Processing Systems 32, NeurIPS 2019}}. \bibinfo{publisher}{Curran Associates,
  Inc.}, \bibinfo{pages}{3518--3531}.
\newblock


\bibitem[\protect\citeauthoryear{Golatkar, Achille, and Soatto}{Golatkar
  et~al\mbox{.}}{2020}]%
        {Golatkar_2020_CVPR}
\bibfield{author}{\bibinfo{person}{Aditya Golatkar},
  \bibinfo{person}{Alessandro Achille}, {and} \bibinfo{person}{Stefano
  Soatto}.} \bibinfo{year}{2020}\natexlab{}.
\newblock \showarticletitle{Eternal Sunshine of the Spotless Net: Selective
  Forgetting in Deep Networks}. In \bibinfo{booktitle}{{\em IEEE/CVF Conference
  on Computer Vision and Pattern Recognition, CVPR 2020}}.
  \bibinfo{publisher}{{IEEE}}, \bibinfo{pages}{9301--9309}.
\newblock


\bibitem[\protect\citeauthoryear{Guo, Goldstein, Hannun, and van~der
  Maaten}{Guo et~al\mbox{.}}{2020}]%
        {guo2019certified}
\bibfield{author}{\bibinfo{person}{Chuan Guo}, \bibinfo{person}{Tom Goldstein},
  \bibinfo{person}{Awni Hannun}, {and} \bibinfo{person}{Laurens van~der
  Maaten}.} \bibinfo{year}{2020}\natexlab{}.
\newblock \showarticletitle{Certified Data Removal from Machine Learning
  Models}. In \bibinfo{booktitle}{{\em 37th International Conference on Machine
  Learning, ICML 2020}}. \bibinfo{publisher}{PMLR}.
\newblock
\newblock
\shownote{To appear.}


\bibitem[\protect\citeauthoryear{Hochreiter and Schmidhuber}{Hochreiter and
  Schmidhuber}{1997}]%
        {Hochreiter97}
\bibfield{author}{\bibinfo{person}{Sepp Hochreiter} {and}
  \bibinfo{person}{J{\"{u}}rgen Schmidhuber}.} \bibinfo{year}{1997}\natexlab{}.
\newblock \showarticletitle{Long Short-Term Memory}.
\newblock \bibinfo{journal}{{\em Neural Computation\/}} \bibinfo{volume}{9},
  \bibinfo{number}{8} (\bibinfo{year}{1997}), \bibinfo{pages}{1735--1780}.
\newblock


\bibitem[\protect\citeauthoryear{Intel}{Intel}{2020}]%
        {SGX}
\bibfield{author}{\bibinfo{person}{Intel}.} \bibinfo{year}{2020}\natexlab{}.
\newblock \bibinfo{title}{{Software Guard Extensions (SGX)}}.
\newblock   (\bibinfo{year}{2020}).
\newblock
\showURL{%
\url{https://software.intel.com/en-us/sgx}}


\bibitem[\protect\citeauthoryear{Lang}{Lang}{1995}]%
        {Lang:ICML95}
\bibfield{author}{\bibinfo{person}{Ken Lang}.} \bibinfo{year}{1995}\natexlab{}.
\newblock \showarticletitle{News{W}eeder: Learning to Filter Netnews}. In
  \bibinfo{booktitle}{{\em 12th International Machine Learning Conference on
  Machine Learning, ICML 1995}}. \bibinfo{publisher}{Morgan Kaufmann},
  \bibinfo{pages}{331--339}.
\newblock


\bibitem[\protect\citeauthoryear{Marcus, Santorini, and Marcinkiewicz}{Marcus
  et~al\mbox{.}}{1993}]%
        {marcus1993building}
\bibfield{author}{\bibinfo{person}{Mitchell~P. Marcus},
  \bibinfo{person}{Beatrice Santorini}, {and} \bibinfo{person}{Mary~Ann
  Marcinkiewicz}.} \bibinfo{year}{1993}\natexlab{}.
\newblock \showarticletitle{Building a Large Annotated Corpus of {E}nglish: The
  {P}enn {T}reebank}.
\newblock \bibinfo{journal}{{\em Computational Linguistics\/}}
  \bibinfo{volume}{19}, \bibinfo{number}{2} (\bibinfo{year}{1993}),
  \bibinfo{pages}{313--330}.
\newblock


\bibitem[\protect\citeauthoryear{McMahan, Ramage, Talwar, and Zhang}{McMahan
  et~al\mbox{.}}{2018}]%
        {DBLP:conf/iclr/McMahanRT018}
\bibfield{author}{\bibinfo{person}{H.~Brendan McMahan}, \bibinfo{person}{Daniel
  Ramage}, \bibinfo{person}{Kunal Talwar}, {and} \bibinfo{person}{Li Zhang}.}
  \bibinfo{year}{2018}\natexlab{}.
\newblock \showarticletitle{Learning Differentially Private Recurrent Language
  Models}. In \bibinfo{booktitle}{{\em 6th International Conference on Learning
  Representations, ICLR 2018}}. \bibinfo{publisher}{OpenReview.net}.
\newblock


\bibitem[\protect\citeauthoryear{Merity, Xiong, Bradbury, and Socher}{Merity
  et~al\mbox{.}}{2017}]%
        {merity2016pointer}
\bibfield{author}{\bibinfo{person}{Stephen Merity}, \bibinfo{person}{Caiming
  Xiong}, \bibinfo{person}{James Bradbury}, {and} \bibinfo{person}{Richard
  Socher}.} \bibinfo{year}{2017}\natexlab{}.
\newblock \showarticletitle{Pointer Sentinel Mixture Models}. In
  \bibinfo{booktitle}{{\em 5th International Conference on Learning
  Representations, ICLR 2017}}. \bibinfo{publisher}{OpenReview.net}.
\newblock


\bibitem[\protect\citeauthoryear{Radford, Wu, Child, Luan, Amodei, and
  Sutskever}{Radford et~al\mbox{.}}{2019}]%
        {radford2019language}
\bibfield{author}{\bibinfo{person}{Alec Radford}, \bibinfo{person}{Jeff Wu},
  \bibinfo{person}{Rewon Child}, \bibinfo{person}{David Luan},
  \bibinfo{person}{Dario Amodei}, {and} \bibinfo{person}{Ilya Sutskever}.}
  \bibinfo{year}{2019}\natexlab{}.
\newblock \bibinfo{booktitle}{{\em Language Models are Unsupervised Multitask
  Learners}}.
\newblock \bibinfo{type}{{T}echnical {R}eport}. \bibinfo{institution}{OpenAI}.
\newblock


\bibitem[\protect\citeauthoryear{Salem, Bhattacharyya, Backes, Fritz, and
  Zhang}{Salem et~al\mbox{.}}{2019a}]%
        {DBLP:journals/corr/abs-1904-01067}
\bibfield{author}{\bibinfo{person}{Ahmed Salem}, \bibinfo{person}{Apratim
  Bhattacharyya}, \bibinfo{person}{Michael Backes}, \bibinfo{person}{Mario
  Fritz}, {and} \bibinfo{person}{Yang Zhang}.}
  \bibinfo{year}{2019}\natexlab{a}.
\newblock \bibinfo{title}{Updates-Leak: Data Set Inference and Reconstruction
  Attacks in Online Learning}.
\newblock   (\bibinfo{year}{2019}).
\newblock
\showeprint[arxiv]{cs.CR/1904.01067}


\bibitem[\protect\citeauthoryear{Salem, Zhang, Humbert, Berrang, Fritz, and
  Backes}{Salem et~al\mbox{.}}{2019b}]%
        {Salem:NDSS19}
\bibfield{author}{\bibinfo{person}{Ahmed Salem}, \bibinfo{person}{Yang Zhang},
  \bibinfo{person}{Mathias Humbert}, \bibinfo{person}{Pascal Berrang},
  \bibinfo{person}{Mario Fritz}, {and} \bibinfo{person}{Michael Backes}.}
  \bibinfo{year}{2019}\natexlab{b}.
\newblock \showarticletitle{{ML}-Leaks: Model and Data Independent Membership
  Inference Attacks and Defenses on Machine Learning Models}. In
  \bibinfo{booktitle}{{\em 26th Annual Network and Distributed System Security
  Symposium, NDSS 2019}}. \bibinfo{publisher}{The Internet Society}.
\newblock


\bibitem[\protect\citeauthoryear{Shokri, Stronati, Song, and Shmatikov}{Shokri
  et~al\mbox{.}}{2017}]%
        {DBLP:conf/sp/ShokriSSS17}
\bibfield{author}{\bibinfo{person}{Reza Shokri}, \bibinfo{person}{Marco
  Stronati}, \bibinfo{person}{Congzheng Song}, {and} \bibinfo{person}{Vitaly
  Shmatikov}.} \bibinfo{year}{2017}\natexlab{}.
\newblock \showarticletitle{Membership Inference Attacks Against Machine
  Learning Models}. In \bibinfo{booktitle}{{\em 38th {IEEE} Symposium on
  Security and Privacy, {S\&P} 2017}}. \bibinfo{publisher}{IEEE Computer
  Society}, \bibinfo{pages}{3--18}.
\newblock


\bibitem[\protect\citeauthoryear{Song and Shmatikov}{Song and
  Shmatikov}{2019}]%
        {natural-auditor}
\bibfield{author}{\bibinfo{person}{Congzheng Song} {and}
  \bibinfo{person}{Vitaly Shmatikov}.} \bibinfo{year}{2019}\natexlab{}.
\newblock \showarticletitle{Auditing Data Provenance in Text-Generation
  Models}. In \bibinfo{booktitle}{{\em Proceedings of the 25th ACM SIGKDD
  International Conference on Knowledge Discovery \& Data Mining, KDD 2019}}.
  \bibinfo{publisher}{ACM}, \bibinfo{pages}{196–206}.
\newblock


\bibitem[\protect\citeauthoryear{{Song}, {Chaudhuri}, and {Sarwate}}{{Song}
  et~al\mbox{.}}{2013}]%
        {6736861}
\bibfield{author}{\bibinfo{person}{S. {Song}}, \bibinfo{person}{K.
  {Chaudhuri}}, {and} \bibinfo{person}{A.~D. {Sarwate}}.}
  \bibinfo{year}{2013}\natexlab{}.
\newblock \showarticletitle{Stochastic Gradient Descent with Differentially
  Private Updates}. In \bibinfo{booktitle}{{\em 1st IEEE Global Conference on
  Signal and Information Processing, GlobalSIP 2013}}. \bibinfo{publisher}{IEEE
  Computer Society}, \bibinfo{pages}{245--248}.
\newblock


\bibitem[\protect\citeauthoryear{Union}{Union}{2016}]%
        {GDPR}
\bibfield{author}{\bibinfo{person}{European Union}.}
  \bibinfo{year}{2016}\natexlab{}.
\newblock \bibinfo{title}{Regulation (EU) 2016/679 of the European Parliament
  and of the Council of 27 April 2016 on the protection of natural persons with
  regard to the processing of personal data and on the free movement of such
  data, and repealing Directive 95/46/EC (General Data Protection Regulation)}.
\newblock   (\bibinfo{year}{2016}).
\newblock


\bibitem[\protect\citeauthoryear{Vaswani, Shazeer, Parmar, Uszkoreit, Jones,
  Gomez, Kaiser, and Polosukhin}{Vaswani et~al\mbox{.}}{2017}]%
        {vaswani2017attention}
\bibfield{author}{\bibinfo{person}{Ashish Vaswani}, \bibinfo{person}{Noam
  Shazeer}, \bibinfo{person}{Niki Parmar}, \bibinfo{person}{Jakob Uszkoreit},
  \bibinfo{person}{Llion Jones}, \bibinfo{person}{Aidan~N. Gomez},
  \bibinfo{person}{{\L}ukasz Kaiser}, {and} \bibinfo{person}{Illia
  Polosukhin}.} \bibinfo{year}{2017}\natexlab{}.
\newblock \showarticletitle{Attention is All You Need}. In
  \bibinfo{booktitle}{{\em Advances in Neural Information Processing Systems
  30, NIPS 2017}}. \bibinfo{publisher}{Curran Associates, Inc.},
  \bibinfo{pages}{5998--6008}.
\newblock


\bibitem[\protect\citeauthoryear{Vijayakumar, Cogswell, Selvaraju, Sun, Lee,
  Crandall, and Batra}{Vijayakumar et~al\mbox{.}}{2018}]%
        {VijayakumarCSSL18}
\bibfield{author}{\bibinfo{person}{Ashwin~K. Vijayakumar},
  \bibinfo{person}{Michael Cogswell}, \bibinfo{person}{Ramprasaath~R.
  Selvaraju}, \bibinfo{person}{Qing Sun}, \bibinfo{person}{Stefan Lee},
  \bibinfo{person}{David~J. Crandall}, {and} \bibinfo{person}{Dhruv Batra}.}
  \bibinfo{year}{2018}\natexlab{}.
\newblock \showarticletitle{Diverse Beam Search for Improved Description of
  Complex Scenes}. In \bibinfo{booktitle}{{\em 32nd {AAAI} Conference on
  Artificial Intelligence, AAAI 2018}}. \bibinfo{publisher}{{AAAI} Press},
  \bibinfo{pages}{7371--7379}.
\newblock


\bibitem[\protect\citeauthoryear{Yeom, Giacomelli, Fredrikson, and Jha}{Yeom
  et~al\mbox{.}}{2018}]%
        {YeomGFJ18}
\bibfield{author}{\bibinfo{person}{Samuel Yeom}, \bibinfo{person}{Irene
  Giacomelli}, \bibinfo{person}{Matt Fredrikson}, {and} \bibinfo{person}{Somesh
  Jha}.} \bibinfo{year}{2018}\natexlab{}.
\newblock \showarticletitle{Privacy Risk in Machine Learning: Analyzing the
  Connection to Overfitting}. In \bibinfo{booktitle}{{\em 31st {IEEE} Computer
  Security Foundations Symposium, {CSF} 2018}}. \bibinfo{publisher}{{IEEE}
  Computer Society}, \bibinfo{pages}{268--282}.
\newblock


\bibitem[\protect\citeauthoryear{Zaremba, Sutskever, and Vinyals}{Zaremba
  et~al\mbox{.}}{2014}]%
        {DBLP:journals/corr/ZarembaSV14}
\bibfield{author}{\bibinfo{person}{Wojciech Zaremba}, \bibinfo{person}{Ilya
  Sutskever}, {and} \bibinfo{person}{Oriol Vinyals}.}
  \bibinfo{year}{2014}\natexlab{}.
\newblock \bibinfo{title}{Recurrent Neural Network Regularization}.
\newblock   (\bibinfo{year}{2014}).
\newblock
\showeprint[arxiv]{cs.NE/1409.2329}


\end{thebibliography}
\bibliographystyle{ACM-Reference-Format}

\appendix

\section{Results for \texttt{talk.politics.mideast}}

Table~\ref{tbl:mideast_results} (deferred from
Section~\ref{sec:res_real_world}) shows the highest-scoring sequences
of length 4 in a group beam search with 5 groups for the
\texttt{talk.politics.mideast} dataset for RNN and Transformer
architectures.

\end{document}